\documentclass[sigconf]{acmart}

\usepackage{amsmath,amsfonts,bm}

\def\eqref#1{equation~\ref{#1}}
\def\1{\bm{1}}

\def\rd{{\textnormal{d}}}

\def\vm{{\bm{m}}}

\def\vx{{\bm{x}}}

\def\vz{{\bm{z}}}

\def\mA{{\bm{A}}}

\def\mX{{\bm{X}}}

\DeclareMathAlphabet{\mathsfit}{\encodingdefault}{\sfdefault}{m}{sl}
\SetMathAlphabet{\mathsfit}{bold}{\encodingdefault}{\sfdefault}{bx}{n}

\def\gE{{\mathcal{E}}}

\def\gG{{\mathcal{G}}}

\def\gM{{\mathcal{M}}}

\def\gV{{\mathcal{V}}}

\def\sR{{\mathbb{R}}}

\def\sX{{\mathbb{X}}}
\def\sY{{\mathbb{Y}}}

\DeclareMathOperator*{\argmax}{arg\,max}
\DeclareMathOperator*{\argmin}{arg\,min}

\usepackage{soul}
\usepackage{url}
\usepackage[utf8]{inputenc}
\usepackage{graphicx}
\usepackage{amsmath}
\usepackage{amsthm}
\usepackage{booktabs}

\usepackage{amsmath,amssymb,amsfonts,amssymb,multirow}
\usepackage{microtype}
\usepackage{graphicx}
\usepackage{subfigure}

\usepackage{scalerel}
\usepackage{afterpage}
\usepackage{nicefrac}       % compact symbols for 1/2, etc.
\usepackage{microtype}      % microtypography
\usepackage{xcolor}         % colors
\usepackage{multirow}
\usepackage{enumitem}
\usepackage{bm}
\usepackage{graphics}
\usepackage{wrapfig}
\usepackage{pifont}
\usepackage{balance}

\newcommand{\whiteding}[1]{\ding{\numexpr171+#1\relax}}

\usepackage{amsthm}

\usepackage{thmtools}
\usepackage{thm-restate}

\usepackage{wrapfig}
\usepackage{multirow}
\usepackage{makecell}
\usepackage{mathtools}
\usepackage{array, arydshln}
\usepackage{amssymb}
\usepackage{wrapfig}
\usepackage{diagbox}
\usepackage[boxed,ruled,vlined,linesnumbered]{algorithm2e}

\SetCommentSty{mycommfont}
\newcolumntype{P}[1]{>{\centering\arraybackslash}p{#1}}

\usepackage{tabu}

\newcolumntype{L}{>{$}l<{$}}
\newcolumntype{C}{>{$}c<{$}}
\newcolumntype{R}{>{$}r<{$}}

\newcommand{\ie}{i.e.,~}

\AtBeginDocument{%
 }

\copyrightyear{2023}
\acmYear{2023}
\setcopyright{rightsretained}
\acmConference[CIKM '23]{Proceedings of the 32nd ACM International Conference on Information and Knowledge Management}{October 21--25, 2023}{Birmingham, United Kingdom}
\acmBooktitle{Proceedings of the 32nd ACM International Conference on Information and Knowledge Management (CIKM '23), October 21--25, 2023, Birmingham, United Kingdom}\acmDOI{10.1145/3583780.3614778}
\acmISBN{979-8-4007-0124-5/23/10}

\makeatletter
\gdef\@copyrightpermission{
  \begin{minipage}{0.3\columnwidth}
   \href{https://creativecommons.org/licenses/by/4.0/}{\includegraphics[width=0.90\textwidth]{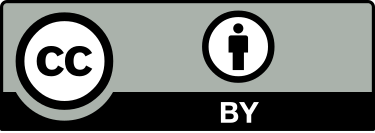}}
  \end{minipage}\hfill
  \begin{minipage}{0.7\columnwidth}
   \href{https://creativecommons.org/licenses/by/4.0/}{This work is licensed under a Creative Commons Attribution International 4.0 License.}
  \end{minipage}
  \vspace{5pt}
}
\makeatother

\begin{document}

\title{AKE-GNN: Effective Graph Learning with \\Adaptive Knowledge Exchange}

\renewcommand{\shorttitle}{AKE-GNN: Effective Graph Learning with Adaptive Knowledge Exchange}

\author{Liang Zeng}
\authornote{The first two authors contribute equally to this work.}
\affiliation{
  \country{Institute for Interdisciplinary Information Sciences (IIIS), Tsinghua University, Beijing, China}
}
\email{zengl18@mails.tsinghua.edu.cn}

\author{Jin Xu}
\authornotemark[1]
\affiliation{
  \country{Institute for Interdisciplinary Information Sciences (IIIS), Tsinghua University, Beijing, China}
}
\email{jxu3425@gmail.com}

\author{Zijun Yao}
\affiliation{
  \country{Department of Computer Science and Technology, Tsinghua University, Beijing, China}
}
\email{yaozj20@mails.tsinghua.edu.cn}

\author{Yanqiao Zhu}
\affiliation{
  \country{Department of Computer Science
University of California, Los Angeles, CA, USA}
}
\email{yzhu@cs.ucla.edu}

\author{Jian Li}
\affiliation{
  \country{Institute for Interdisciplinary Information Sciences (IIIS), Tsinghua University, Beijing, China}
}
\email{lijian83@mail.tsinghua.edu.cn}

\begin{abstract}
    Graph Neural Networks~(GNNs) have already been widely used in various graph mining tasks. However, recent works reveal that the learned weights~(channels) in well-trained GNNs are highly redundant, which inevitably limits the performance of GNNs. Instead of removing these redundant channels for efficiency consideration, we aim to reactivate them to enlarge the representation capacity of GNNs for effective graph learning.
In this paper, we propose to substitute these redundant channels with other informative channels to achieve this goal.
We introduce a novel GNN learning framework named AKE-GNN, which performs the \underline{A}daptive \underline{K}nowledge \underline{E}xchange strategy among multiple graph views generated by graph augmentations. 
AKE-GNN first trains multiple GNNs each corresponding to one graph view to obtain informative channels. Then, AKE-GNN iteratively exchanges redundant channels in the weight parameter matrix of one GNN with informative channels of another GNN in a layer-wise manner. 
Additionally, existing GNNs can be seamlessly incorporated into our framework. AKE-GNN achieves superior performance compared with various baselines across a suite of experiments on node classification, link prediction, and graph classification. In particular, we conduct a series of experiments on 15 public benchmark datasets, 8 popular GNN models, and 3 graph tasks and show that AKE-GNN consistently outperforms existing popular GNN models and even their ensembles. Extensive ablation studies and analyses on knowledge exchange methods validate the effectiveness of AKE-GNN.
% Adaptively exchanges diverse knowledge learned from
\end{abstract}

\begin{CCSXML}
<ccs2012>
   <concept>      <concept_id>10010147.10010257.10010293.10010294</concept_id>
       <concept_desc>Computing methodologies~Neural networks</concept_desc>
       <concept_significance>500</concept_significance>
       </concept>
   <concept>
       <concept_id>10010147.10010257.10010293.10010319</concept_id>
       <concept_desc>Computing methodologies~Learning latent representations</concept_desc>
       <concept_significance>500</concept_significance>
       </concept>
   <concept>
       <concept_id>10002951.10003227.10003351</concept_id>
       <concept_desc>Information systems~Data mining</concept_desc>
       <concept_significance>500</concept_significance>
       </concept>
 </ccs2012>
\end{CCSXML}

\ccsdesc[500]{Computing methodologies~Neural networks}
\ccsdesc[500]{Computing methodologies~Learning latent representations}
\ccsdesc[500]{Information systems~Data mining}

\keywords{Graph neural networks, Graph representation learning, Adaptive knowledge exchange}

\maketitle

\section{Introduction}
\begin{figure*}[t]
    \centering
    \includegraphics[width=16cm]{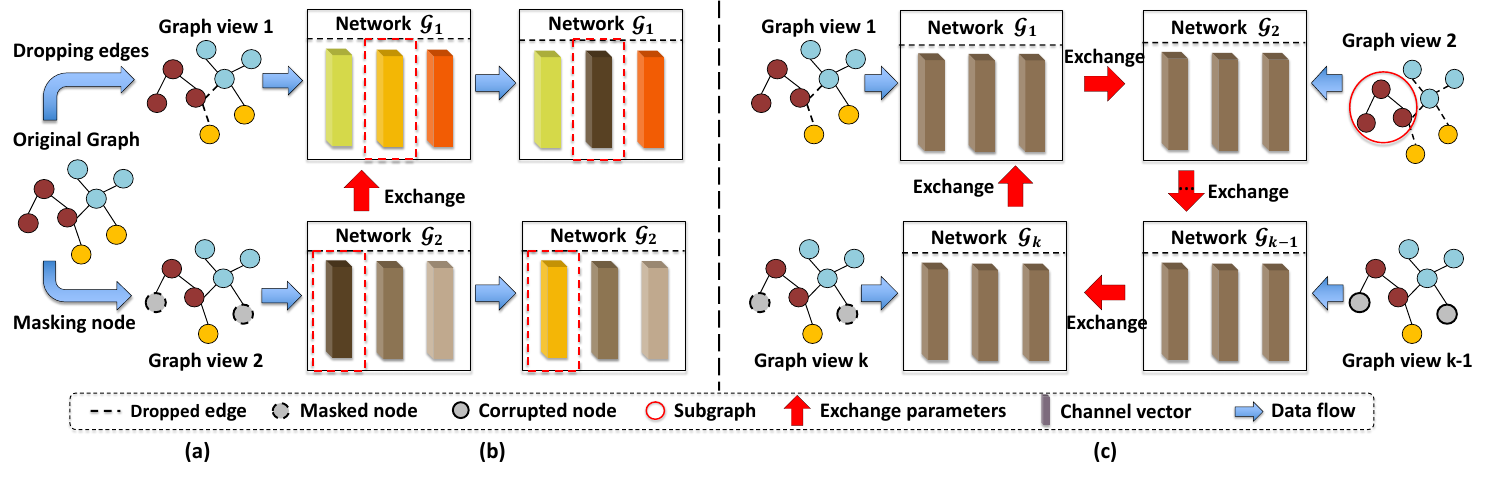}
    \caption{The illustrative schematic diagram of the proposed AKE-GNN framework. (a) Two generated graph views~(\ie masking node features and dropping edges). (b) Adaptive knowledge exchange by exchanging channel-wise parameters among two graph views (in one layer for illustration). (c) AKE-GNN in the multiple GNN case~(best viewed in color).}
    \label{fig:framework}
\end{figure*}
Graph Neural Networks~(GNNs), as the powerful tool for modeling relational inductive bias~\cite{battaglia2018relational,barabasi2013network} to jointly encode graph structure and node features of the input graph~\cite{hamilton2020graph}, have been widely employed for analyzing graph-mining tasks, including node classification~\cite{kipf2016semi,velivckovic2017graph,hamilton2017inductive,chen2020simple,wu2019simplifying}, link prediction~\cite{zhang2018link,ying2018graph}, and graph classification~\cite{xu2018powerful,errica2019fair}.
% Typically, the training process of GNN models follows the message-passing scheme to learn effective node/graph representations that capture structural information of a given graph~\cite{kipf2016semi,velivckovic2017graph,xu2018powerful}.
Despite the prevalence and effectiveness of GNN models, as discussed in recent works~\cite{chen2021unified,jin2022feature}, there exist redundant channels of the weight parameter matrix in a well-trained GNN model. These redundant channels can be removed without performance degradation.
Existing works mainly remove these redundant channels from the perspective of efficiency. 
However, non-structured channel pruning methods are not hardware-friendly~\cite{han2015learning} and thus suffer from limited efficiency improvement~\cite{li2016pruning}. 
Moreover, these methods often improve efficiency of the model with a slight sacrifice of effectiveness~\cite{chen2021unified,frankle2018lottery}. 
Therefore, from a novel and practical perspective of effectiveness, we propose to substitute these redundant channels with informative channels to enrich knowledge of GNN models for effective graph learning.
To achieve this goal, we need to tackle two unique technical challenges: 1) How to obtain informative channels? 2) How to exchange redundant channels with informative channels effectively?

For obtaining informative channels, we are inspired by recent advances in multi-view GNNs, whose multiple graph views generated by graph augmentations can provide complementary information of a graph from different aspects. Most GNN models are trained in an end-to-end supervised manner to learn effective node/graph representations in a single-view graph.
As argued in some recent works~\cite{wang2020gcn,xie2020mgat,xu2019multi}, such training methods can only capture partial information from the complex input, hence may not generalize well on unseen nodes/graphs.
As a result, researchers propose new training algorithms that generate multiple views from the input graph and then build multi-view GNNs~\cite{wang2020gcn,cheng2020multi}. The idea is that each view captures knowledge from one certain aspect, and knowledge learned from different views is fused to enhance node/graph representation. Representative models include AM-GCN \cite{wang2020gcn} and MGAT \cite{xie2020mgat} which utilize multi-head attention modules to fuse feature and topology knowledge, and MAGCN~\cite{cheng2020multi} which develops multi-view attribute graph convolution encoders.
% , and MTNE~\cite{xu2019multi}, in which knowledge is propagated via the so-called knowledge sharing embedding vectors.

For exchanging channels effectively, we propose a novel GNN learning framework, called \underline{A}daptive \underline{K}nowledge \underline{E}xchange GNNs (AKE-GNN), which fuses diverse knowledge learned from multiple graph views generated by graph augmentations. AKE-GNN adaptively exchanges parameters among those graph views.
% , by adaptively exchanging parameters among those views.
The advantage of AKE-GNN is that we do not need to modify the model architecture or training loss functions~\cite{jin2022feature}, and thus existing GNN models can be seamlessly incorporated into our framework.

AKE-GNN contains two training phases: an individual learning phase and a knowledge exchange phase. In the individual learning phase, we construct multiple views by stochastic graph augmentation functions~\cite{zhu2020graph}, and GNNs sharing the same backbone model learn those graph views independently to obtain informative channels. In the knowledge exchange phase, we design a channel-wise adaptive exchange method that repeatedly replaces redundant channels in one GNN with the informative channels from another GNN in a layer-wise manner. 
Furthermore, we show the extension of AKE-GNN to more than two graph views. 
Comprehensive experiments show that AKE-GNN consistently achieves superior performance over existing popular GNN models and their ensembles on representative graph tasks including node classification, link predictions, and graph classification, and across various domains including bioinformatics~(e.g., to predict the property of the protein) and social networks~(e.g., to predict the co-authorship).
In shot, our main \textbf{contributions} are: 
% summarized as follows:
\begin{itemize}[leftmargin=10pt]
    \item We present a novel GNN learning framework, namely AKE-GNN, which adaptively exchanges knowledge from multiple GNNs learned on diverse graph views for effective graph learning.
    \item Existing backbone GNN models can be seamlessly incorporated into AKE-GNN without modifying the original configurations such as the learning rate and the number of layers. Moreover, AKE-GNN introduces no extra computational overheads to the inference stage. 
    \item We extensively evaluate the effectiveness of AKE-GNN on 15 public datasets, 8 popular GNN models, and 3 graph tasks. AKE-GNN consistently outperforms corresponding GNN backbone models by an average of 1.9\%$\sim$3.9\% in terms of absolute accuracy improvements and even their ensembles. 
    In addition, extensive ablation studies and analyses on the proposed knowledge exchange method also validate the effectiveness of AKE-GNN.
\end{itemize}
\section{Related Work}
\paragraph{Joint learning of multiple graph views.} Multi-view joint learning aims to jointly model~(generated) multiple graph views to improve the generalization performance~\cite{wang2020gcn,xie2020mgat,ma2020multi,xu2019multi,cheng2020multi}. Most existing works leverage graph augmentations to generate multiple views from the original graph, and then design specific architectures to collaboratively fuse knowledge learned from different graph views to enhance their ability of graph representation learning.  
AM-GCN~\cite{wang2020gcn} explicitly constructs the node feature graph view and the topology graph view, and then employs two GNNs with attention mechanisms to extract knowledge from these two aspects.
MGAT~\cite{xie2020mgat} automatically generates multiple views via graph augmentations and then designs an attention-based architecture to collaboratively integrate multiple types of knowledge in different views.
MAGCN~\cite{cheng2020multi} develops multi-view attribute graph convolution encoders with attention mechanisms for learning graph embeddings from multi-view graph data.
Different from these methods, our method retains the benefits of joint modeling multiple views via an adaptive knowledge exchange framework while not requiring dedicated architecture designs.
Additionally, graph contrastive learning~(GCL) methods~\cite{hassani2020contrastive,you2020graph,qiu2020gcc,zhu2020graph} leverage generated multiple graph views to maximize the feature consistency among these views. However, GCL methods operate within a self-supervised learning regime, where label information is not available during the training phase. In contract, AKE-GNN is grounded in a supervised learning setting to facilitate knowledge exchange of parameters from informative to redundant channels.

\paragraph{Weight re-activating.} 
Our adaptive knowledge exchange framework is conceptually connected to weight re-activating methods.
Grafting~\cite{meng2020filter} improves the network performance by grafting external information~(weights) on the same data source to re-activate invalid filters in computer vision tasks. DeCorr~\cite{jin2022feature} introduces the explicit feature dimension decorrelation term into the loss objective to tackle the feature overcorrelation issue in GNNs.
In contrast, our work aims at fusing different knowledge from GNNs trained on multiple (generated) graph views. 
Since different graph views share different parts of knowledge that should not be repeated in just one GNN, we propose an adaptive approach to iteratively exchange complementary knowledge from different graph views for more effective graph learning.
\section{AKE-GNN: The Proposed Framework}
\label{sec:method}
\begin{table}[t]
\caption{Results of accuracy~(\%) with standard deviations of GCN by successively removing redundant channel pairs based on 50 repeated runs. The hidden size~(channel) of GCN is 16.}
\small
\centering
\label{table:1}
\scalebox{1.}{
\setlength{\tabcolsep}{4pt}
\begin{tabular}{ccccccccc}
\toprule
\begin{tabular}{c}\#Remaining\\channels\end{tabular} & 16        & 14        & 12        & 10   & 8     & 6     & 4     & 2      \\ \midrule
\multirow{2}*{Accuracy}  & $81.4$ & $81.5$ & $81.0$ & $80.2$ & $79.7$ & $75.6$ & $73.4$ & $51.9$ \\ \cmidrule{2-9}
& ${\pm0.6}$ & ${\pm0.7}$ & ${\pm0.7}$ & ${\pm1.2}$ & ${\pm1.3}$ & ${\pm1.5}$ & ${\pm5.1}$ & ${\pm7.5}$ \\
\bottomrule
\end{tabular}
}
\end{table}

In this section, we first present a preliminary study to investigate the redundancy issue of the weight parameter matrix in GNNs~(Sec.~\ref{sec:redundancy}). Then we introduce the AKE-GNN framework on two graph views with its two training phases~(Sec.~\ref{sec:individual-phase}~\&~\ref{sec:evolution-phase}). 
% an individual learning phase~(Sec.~\ref{sec:individual-phase}) and a knowledge exchange phase (Sec.~\ref{sec:evolution-phase}). 
We finally extend AKE-GNN to the multiple GNN case~(Sec.~\ref{sec:multiple-gnn}). 
The overall framework of AKE-GNN is shown in Fig.~\ref{fig:framework}.

\paragraph{Notations.}
\label{sec:preliminary}
Let $\gG=\left(\gV, \gE, \mX\right)$ denote a graph, where $\gV$ is a set of $N$ nodes, and $\gE$ is a set of edges between nodes. $\mX = \left[\vx_1, \vx_2, \cdots, \vx_N\right]^{T}$ $\in \sR^{n \times d}$ represents the node feature matrix and $\vx_i \in \sR^d$ is the feature vector of node $v_i$, where $d$ is the number of channels in the feature matrix $\mX$. The adjacency matrix $\mA \in \{0,1\}^{N \times N}$ is defined by $\mA_{i,j}=1$ if $\left(v_i, v_j\right) \in \gE$ and $0$ otherwise.
We denote the (generated) multiple views as $\gG_{\gM} = \{\gG_1, \cdots, \gG_V\}$, where $\gG_v$ is the $v$-th view of the original input graph $\gG$. Note that all of the graph views share the same node set.

\subsection{Redundancy on GNN Models}
\label{sec:redundancy}
\citet{jin2022feature} have empirically found that the weight parameter matrix of GNNs has a high tendency to contain redundant channels resulting from standard GNN training, \ie high Pearson correlation among channels in the weight matrix.
We verified this phenomenon by conducting experiments on Cora with GCN. We successively find a pair of output channels with the highest Pearson correlation in the weight matrix and then prune these two channels and re-train the resultant GCN model, starting from hidden size 16. 
In Table~\ref{table:1}, we find that several channels have minor impacts on the output, and pruning these redundant channels does not degrade the performance.
This preliminary study inspires us that GNN models indeed contain highly correlated channels in the weight matrix, which cannot introduce extra useful information.
It naturally spurs a question: can we further improve the performance of GNN models by adaptively exchanging knowledge contained in their learned weights (channels)?
Herein, we need to tackle two unique technical challenges: 1) How to obtain informative channels~(Sec.~\ref{sec:individual-phase})?
2) How to exchange channels effectively~(Sec.~\ref{sec:evolution-phase})?

\begin{figure}[t]
    \centering
    \includegraphics[width=\linewidth]{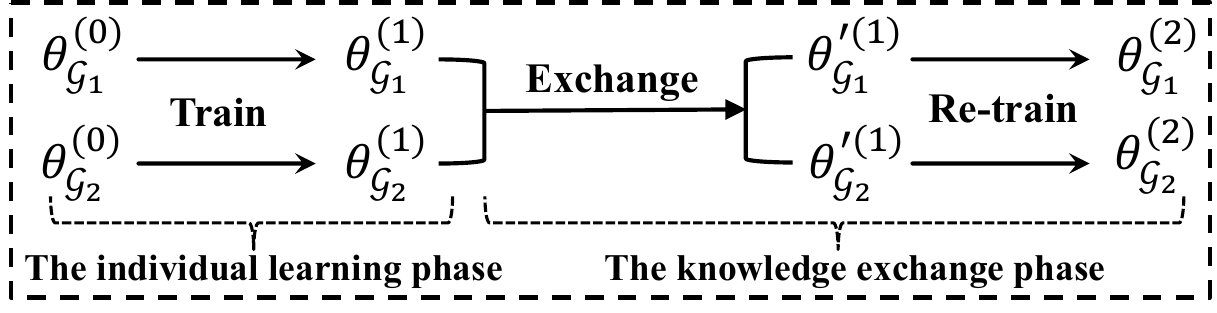}
    % \vspace{-2em}
    \caption{The illustration of the pipeline of parameter updating in AKE-GNN with two GNNs.}
    % \vspace{-1em}
    \label{fig:param-update}
\end{figure}
\subsection{The Individual Learning Phase}
\label{sec:individual-phase}
In this phase, we first generate the multiple graph views by graph augmentations. Then, we train multiple GNNs each corresponding to a generated graph view to obtain informative channels of the weight parameter matrix in GNNs.
% \zl{The intuition is that if the noise in graph views is independent, then the highly correlated shared information might be signal.}
\paragraph{Generating multiple views.} 
To capture different views of the original graph, following previous work~\cite{velivckovic2018deep,zhu2020graph}, we apply stochastic \emph{augmentation functions} to generate multiple views of the original graph and then feed them into GNNs.
Formally, a different view of the original graph $\left(\mX, \mA\right)$ is obtained by $\left(\widetilde{\mX}, \widetilde{\mA}\right) = \mathcal{C} \left(\mX, \mA\right)$, where $\mathcal{C}(\cdot)$ is an \emph{augmentation function}.
We leverage four commonly-used augmentation functions to generate multiple graph views in AKE-GNN~\cite{zhu2020graph,velivckovic2018deep,feng2020graph}, which are \emph{masking node features}, \emph{corrupting node features}, \emph{dropping edges}, and \emph{extracting subgraphs}. 
% The details of graph augmentations can be found in Appendix B.
\begin{itemize}[leftmargin=10pt]
    \item \emph{Masking node features.} Randomly mask a fraction of node attributes with zeros. Formally, the generated matrix of node features $\widetilde{\mX}$ is computed by
    \begin{equation}
        \setcounter{equation}{4}
        \widetilde{\mX} = [\vx_1 \odot \vm, \vx_2 \odot \vm, \cdots, \vx_N \odot \vm],
    \end{equation}
    where $\vm \in \{0,1\}^d$ is a random vector, which is drawn from a Bernoulli distribution, $[ \cdot \; , \cdot]$ is the concatenation operator, and $\odot$ is the element-wise multiplication. \\
    \item \emph{Corrupting node features.} Randomly replace a fraction of node attributes with Gaussian noise. Formally, it can be calculated by
    \begin{equation}
        \widetilde{\mX} = [\vx_1 \odot \vm_1, \vx_2 \odot \vm_2, \cdots, \vx_N \odot \vm_n],
    \end{equation}
    where $\vm_i \in \sR^d$ is a random vector drawn from a Gaussian distribution $\mathcal{N}\left(\mu(\vx_i),1\right)$ independently and $\mu(\cdot)$ denotes the mean value of a vector.
    \item \emph{Dropping edges.} Randomly remove edges in the graph. Formally, we sample a modified subset $\widetilde{\gE}$ from the original edge set $\gE$ with the probability defined as follows:
    \begin{equation}
        P\{\left(u, v\right) \in \widetilde{\gE}\}=1-p_{u v}^{e},
    \end{equation}
    where $\left(u, v\right) \in \gE$ and $p_{u v}^{e}$ is the probability of removing $\left(u, v\right)$.
    \item \emph{Extracting subgraphs.} Extract the induced subgraphs $\gG'= \left(\gV',\gE'\right)$ containing the nodes ang the corresponding edges in a given subset~\cite{velivckovic2018deep}, \ie $\gV' \subseteq \gV$ and $\gE' \subseteq \gE$.
\end{itemize}
Note that AKE-GNN does not require specific graph augmentation functions, and thus other graph augmentation methods can be seamlessly incorporated into our framework. 

\paragraph{Training GNNs.} 
\label{sec:training-of-GNNs}
For any existing GNN model, it can be directly applied in the AKE-GNN framework without modifying its original implementations such as the learning rate and the number of layers.   
We denote a parametrized GNN as $\mathcal{F}_{\bm{\theta}}:  \sX \rightarrow \sY$ with the initial parameter $\bm{\theta}^{\left(0\right)}$, where $\sX$ and $\sY$ are the input space and output space. 
Given $N_{\text{train}}$ paired training data $\{\left(\vx_t, y_t\right)\}_{t=1}^{N_{\text{train}}} \subset \sX \times \sY$, the network $\mathcal{F}_{\bm{\theta}}$ is optimized with a supervised loss $\mathcal{L}$ as follows:
\begin{equation}
    \bm{\theta}^{\left(1\right)} = \argmin_{\bm{\theta}}\mathbb{E}_{\left(\vx,y\right) \in \sX \times \sY} [\mathcal{L}\left(\mathcal{F}_{\bm{\theta}}\left(\vx\right), y\right)] \;, 
    \label{equ:loss}
\end{equation}
where $\bm{\theta}^{\left(1\right)}$ is the parameters of a GNN after optimization.

\begin{algorithm}[t]
    \caption{AKE-GNN with two GNNs}
    \label{alg:gsl}
    \KwIn{Two input graphs $\mathcal{G}_1$ and $\mathcal{G}_2$; two GNNs $\mathcal{F}_{1}$ and $\mathcal{F}_{2}$; $\bm{\bm{\theta}}_{\gG_k}^{l}$ denotes parameters in the $l$-th layer of $\mathcal{F}_{k}$; the number of layers $L$; iteration steps $N$; the number of exchange channels $M$.}
  	$\vartriangleright$ \textbf{The individual learning phase: }~\\
    Update the parameters $\bm{\bm{\theta}}_{\gG_1}$ of GNN $\mathcal{F}_1\left(\mathcal{G}_1;\bm{\bm{\theta}}_{\gG_1}\right)$ and $\bm{\bm{\theta}}_{\gG_2}$ of GNN $\mathcal{F}_2\left(\mathcal{G}_2;\bm{\bm{\theta}}_{\gG_2}\right)$ 
    with Eq.~\ref{equ:loss}.~\\
  	$\vartriangleright$ \textbf{The knowledge exchange phase: }~\\
    \For{$n = \{1,\dots,N\}$}{
        Obtain the source and target GNN: $s = (n-1)\%2+1$, $t = n\%2+1$. \tcc*{for two GNNs, $s,t \in \{0, 1\}$.}
        \For{$l = \{1,\dots,L\}$, $m = \{1,\dots,M\}$ \textbf{parallel}}{
                Calculate the Pearson correlation among all possible pairs of the output channels in $\bm{\bm{\theta}}_{\gG_t}^l$.~\\
                Find a pair of output channels indexed by $idx_1$ and $idx_2$ with the highest correlation.~\\
                Obtain the informative channel $i$ of the source network and the redundant channel $r$ of the target network with Eq.~\ref{eq:informative}.\\ 
                Exchange parameters between two output channels $\bm{\bm{\theta}}_{\gG_s}^{l,i}$ and $\bm{\bm{\theta}}_{\gG_t}^{l,r}$.~\\
        }
    }
    \textbf{Output:} re-trained $\mathcal{F}_{1}$ and $\mathcal{F}_{2}$ according to Eq.~\ref{equ:loss}.
\end{algorithm}
\subsection{The Knowledge Exchange Phase}
\label{sec:evolution-phase}
Earlier works identify that knowledge is contained in the updated parameter values of a neural network~\cite{hinton2015distilling}.
After the individual learning phase, GNNs trained with multiple views have learned knowledge stored in their updated parameters and can take a further step to interact with each other for knowledge exchange.
In the knowledge exchange phase, we take $\bm{\theta}_{\gG_1}^{\left(1\right)}$ and $\bm{\theta}_{\gG_2}^{\left(1\right)}$ of the two GNNs as input, exchange knowledge, and produce $\bm{\theta}_{\gG_1}^{'\left(1\right)}$ and $\bm{\theta}_{\gG_2}^{'\left(1\right)}$ as output, where $\gG_1$ and $\gG_2$ are two corresponding graph views.
Then, we re-train the output parameters $\bm{\theta}_{\gG_1}^{'\left(1\right)}$ and $\bm{\theta}_{\gG_2}^{'\left(1\right)}$ and obtain the final parameters $\bm{\theta}_{\gG_1}^{\left(2\right)}$ and $\bm{\theta}_{\gG_2}^{\left(2\right)}$.
The illustration of the pipeline of parameter updating in AKE-GNN is shown in Fig.~\ref{fig:param-update}.
To exchange knowledge among multiple GNNs, we need to answer the following two questions: 1) How to measure information (knowledge) inside the parameters~(connection weights)? 2) How to adaptively perform knowledge exchange among multiple GNNs?

\paragraph{Entropy.} We leverage entropy to measure information in one layer of a well-trained GNN. As suggested in~\cite{meng2020filter}, the higher entropy the weight matrix has, the more variation~(the less redundant information) the model contains, and then the potentially better performance of the final prediction.
Let $\bm{\bm{\theta}}_{\gG_k}^l \in \sR^{d_i \times d_{i+1}}$ denote the parameters of the $l$-th layer in the corresponding GNN whose input is $\gG_k$, where $d_i$ is the number of channels in the $l$-th layer. 
Following~\cite{meng2020filter,cheng2019utilizing},
we calculate entropy by dividing the range of values in $\bm{\bm{\theta}}_{\gG_k}^l$ into $B$ different bins. Denote the number of values that fall into the $b$-th bin as $n_b$. We use $p_b \triangleq \frac{n_b}{\left(n_1 + \cdots + n_m\right)}$ to approximate the probability of the $b$-th bin, where $n_1 + \cdots + n_B = d_i \times d_{i+1}$. Then, the entropy of $\bm{\bm{\theta}}_{\gG_k}^l$ can be calculated as follows:
\begin{equation}
    H\left(\bm{\bm{\theta}}_{\gG_k}^l\right) = -\sum^{B}_{b=1} p_b \; \text{log}p_b  \;.
    \label{equ:entropy}
\end{equation}
A larger value of $H\left(\bm{\bm{\theta}}_{\gG_k}^l\right)$ usually indicates richer information in the parameters of the $l$-th layer in the corresponding GNN whose input is $\gG_k$, and vice versa. For example, if each element of $\bm{\bm{\theta}}_{\gG_k}^l$ takes the same value~(entropy is 0), $\bm{\bm{\theta}}_{\gG_k}^l$ cannot discriminate which part of the input is more important.

\paragraph{Adaptive exchange.} 
Given quantitative measurements of information in each layer of a GNN, we then consider how to adaptively exchange information among multiple GNNs. 
Since GNNs follow the message passing scheme to iteratively aggregate information from neighbor nodes, the $k$-th layer makes use of the subtree structures of height $k$ rooted at every node. Thus, we only exchange information of the same layer to preserve the consistency of information between two GNNs. 
Let parameters of the source and the target GNN be $\bm{\bm{\theta}}_{\gG_s}$ and $\bm{\bm{\theta}}_{\gG_t}$, respectively. 
We denote parameters in the $l$-th GNN layer trained on the $k$-th graph view as $\bm{\bm{\theta}}_{\gG_k}^l = \small[ \bm{\bm{\theta}}_{\gG_k}^{l,1},\cdots, \bm{\bm{\theta}}_{\gG_k}^{l,C_{l+1}} \small] \in \sR^{C_l \times C_{l+1}}$, where the input channel is $C_l$, the output channel is $C_{l+1}$, and $\bm{\bm{\theta}}_{\gG_k}^{l,j} \in \sR^{C_{l}}$ is the $j$-th output channel vector. 
In each exchange step, our target is to adaptively exchange a redundant output channel in $\bm{\bm{\theta}}_{\gG_t}$ with another informative output channel in $\bm{\bm{\theta}}_{\gG_s}$. 
To exchange the redundant channel, we first calculate the values of \emph{Pearson correlation} among all possible output channel pairs in $\bm{\bm{\theta}}_{\gG_t}^{l}$, and then obtain a pair of redundant channels with the highest correlation, \ie $\text{idx}_1$ and $\text{idx}_2$.
We select an output channel from $\bm{\bm{\theta}}_{\gG_s}^{l}$ to substitute $\bm{\bm{\theta}}_{\gG_t}^{l,\text{idx}_1}$ or $\bm{\bm{\theta}}_{\gG_t}^{l,\text{idx}_2}$ with the purpose to maximize entropy of the new weight parameter matrix $\bm{\bm{\theta}}_{\gG_t}^{l}$. Formally, let $f\left(\mA^{j}, \vx\right)$ be the operator to substitute the $j$-th output channel of the matrix $\mA$ with a vector $\vx$. We find the informative output channel $i\in\left[C_{l+1}\right]$ in the source network $\gG_s$ and the redundant output channel $r\in \{\text{idx}_1, \text{idx}_2\}$ in the target network $\gG_t$ at the $l$-th layer as follows:
\begin{equation}
    i, r = \argmax_{\substack{i\in\left[C_{l+1}\right]\\r\in \{\text{idx}_1, \text{idx}_2\}}} H\left( f\left(\bm{\bm{\theta}}_{\gG_t}^{l,r}, \bm{\bm{\theta}}_{\gG_s}^{l,i} \right) \right) \;.
    \label{eq:informative}
\end{equation}
Finally, we exchange parameters between $\bm{\bm{\theta}}_{\gG_s}^{l,i}$ and $\bm{\bm{\theta}}_{\gG_t}^{l,r}$. By repeating the above exchange step for $M$ times, $\bm{\bm{\theta}}_{\gG_t}^l$ can accept the part of information~($M$ channels) from $\bm{\bm{\theta}}_{\gG_s}^l$ while retaining the useful information in the original network $\bm{\bm{\theta}}_{\gG_t}^l$. We illustrate the procedure in Fig.~\ref{fig:framework}~(b), where $\bm{\bm{\theta}}_{\gG_1}$ and $\bm{\bm{\theta}}_{\gG_2}$ perform adaptive channel-wise parameter exchange in one layer as aforementioned. As a result, $\bm{\bm{\theta}}_{\gG_1}$ exchanges the second channel in the weight matrix with the first channel in $\bm{\bm{\theta}}_{\gG_2}$. Through this procedure, both networks contain information from two graph views. Finally, we re-train two GNNs with the same number of epochs as introduced in Sec.~\ref{sec:training-of-GNNs} to obtain the output predictions.
The complete algorithm of AKE-GNN with two GNNs is summarized in Algorithm~\ref{alg:gsl}.

\begin{figure}[t]
    \centering
    \includegraphics[width=\linewidth]{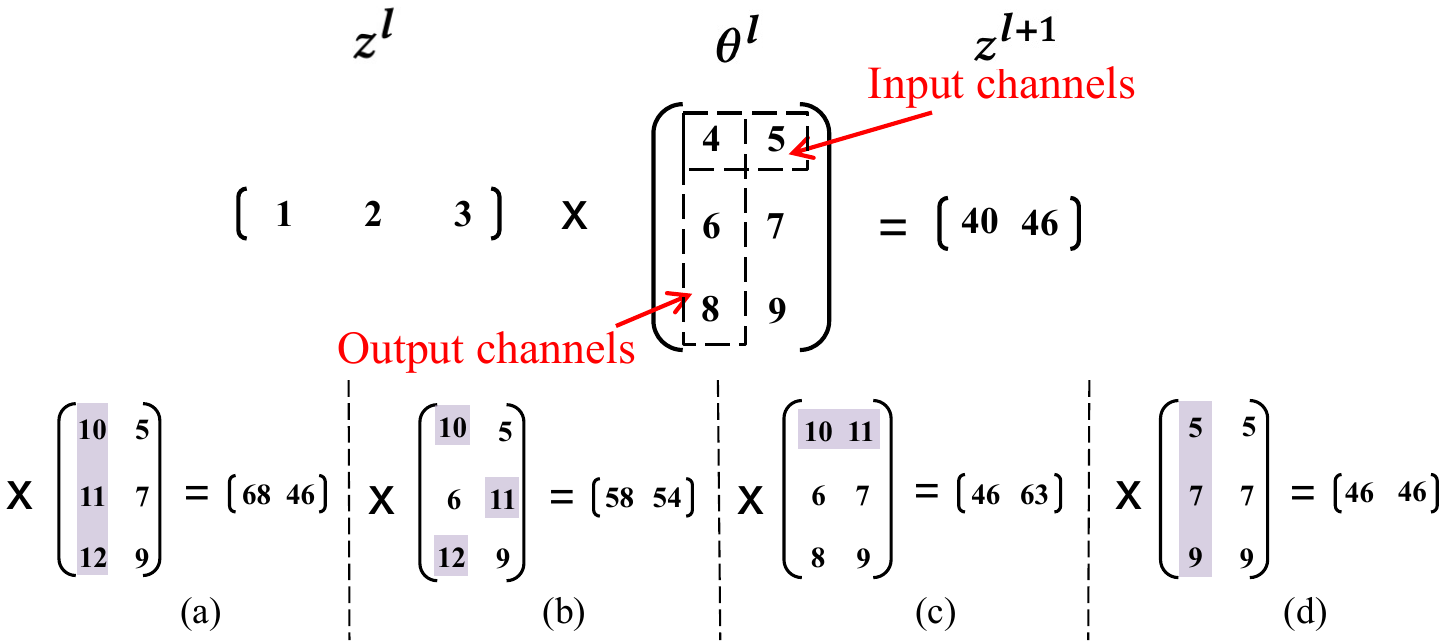}
    % \vspace{-2em}
    \caption{The illustration of exchanging output channels~(a) compared with point-wise exchange~(b), exchanging input channels~(c), and self-exchange output channels~(d). The newly replaced value in the weight matrix is depicted in purple. 
    We only consider features of one node (one row in $\vz^l$) and omit the input feature vector $\vz^l$ in the bottom half of the figure for brevity.}
    % \vspace{-1em}
    \label{fig:tensor-multi}
\end{figure}
\paragraph{Remark.}
To further explain the rationale of our proposed adaptive knowledge exchange method, we present an illustrated example in Fig.~\ref{fig:tensor-multi}.
We denote the input and the output feature in the $l$-th GNN layer as $\vz^l$ and $\vz^{l+1}$, respectively. Let $\theta^l$ be the weight matrix of the $l$-th GNN layer. Each value in $\vz^{l}$/$\vz^{l+1}$ represents the certain feature dimension. 
Each feature dimension in $\vz^{l+1}$ is a function of $\vz^{l}$, which is parameterized by output channels in $\theta^l$.
Thus, exchanging output channels can produce partially modified features in $\vz^{l+1}$.
Comparing Fig.~\ref{fig:tensor-multi}~(a) with (b) and (c), exchanging certain output channels can alter the corresponding features in $\vz^{l+1}$~(e.g., ``40''$\rightarrow$``68'') while keeping other features unchanged~(e.g., ``46''), which explicitly contain information of both the original and the other new network. 
Comparing (a) with (d), self-exchange among output channels in the network itself cannot bring new information and can only obtain repeated features~(e.g., repeated ``46'' in Fig.~\ref{fig:exchange}~(d)). However, exchanging output channels among multiple GNNs can introduce extra information from the other weight matrix and obtain new features~(e.g., ``68'' in Fig.~\ref{fig:exchange}~(a)).
We also present a detailed ablation study on AKE-GNN to compare adaptive channel exchange with other parameter exchange methods such as randomly exchanging parameters (without the adaptive exchange strategy), exchanging parameters with a randomly initialized model (without a well-trained GNN model), and exchanging output channels in the network itself (only disturbing channels) in Sec.~\ref{sec:exchange_method}. 
We empirically find that adaptive knowledge exchange learns more effective graph representations and achieves the highest accuracy 82.8\% on the Cora dataset.

\begin{table}[t]
\caption{Summary of performance on Cora, Citeceer, and Pubmed in terms of accuracy in percentage with standard deviation. ``$\dagger$'' means that we re-implement the adaptive weighting method according to~\protect\citet{meng2020filter}. ``-'' means that the original paper does not report the corresponding results.}
%Best results are marked in bold.
% \vspace{-0.2em}
\label{table:results_of_public}
\small
\centering
\scalebox{1.}{
\setlength{\tabcolsep}{4pt}
\begin{tabular}{cc|ccc}
\Xhline{0.2ex}
\multicolumn{2}{l}{}                        & Cora         & CiteSeer                    & PubMed       \\ \hline
\multirow{11}{*}{\begin{tabular}{c} Single-view \\ GNNs\end{tabular}}  & \multicolumn{1}{|c|}{GCN}                & $81.5$         & $70.3$                        & $79.0$           \\
                             &
                             \multicolumn{1}{|c|}{IncepGCN}             & $83.5$ & $72.7$                & $79.5$ \\
                             &\multicolumn{1}{|c|}{GAT}                  & $83.0_{\pm0.7}$ & $72.5_{\pm0.7}$                & $79.0_{\pm0.3}$ \\
                             & \multicolumn{1}{|c|}{GraphSAGE}            & $78.9_{\pm0.8}$ & $67.4_{\pm0.7}$                & $77.8_{\pm0.6}$ \\
                             & \multicolumn{1}{|c|}{APPNP}                & $83.8_{\pm0.3}$ & $71.6_{\pm0.5}$                & $79.7_{\pm0.3}$ \\
                             & \multicolumn{1}{|c|}{Graph U-Net}          & $84.4_{\pm0.6}$ & $73.2_{\pm0.5}$                & $79.6_{\pm0.2}$ \\
                             & \multicolumn{1}{|c|}{MixHop}              & $81.9_{\pm0.4}$ & $71.4_{\pm0.8}$                & $80.8_{\pm0.6}$ \\
                             & \multicolumn{1}{|c|}{SGC}                  & $82.0_{\pm0.0}$ & $71.9_{\pm0.1}$                & $78.9_{\pm0.0}$ \\
                             & \multicolumn{1}{|c|}{GraphMix}             & $83.9_{\pm0.6}$ & $74.5_{\pm0.6}$                & $81.0_{\pm0.6}$ \\
                             & \multicolumn{1}{|c|}{GCNII}              & $85.5_{\pm0.5}$ & $73.4_{\pm0.6}$                & $80.2_{\pm0.4}$ \\ 
                             & \multicolumn{1}{|c|}{DeCorr}              & $85.3$ & $73.7$                & $80.4$ \\ \hline
\multirow{6}{*}{\begin{tabular}{c} Multi-view \\ GNNs\end{tabular}}  & \multicolumn{1}{|c|}{MAGCN}               & $75.1$         & $71.1$                        & $69.1$         \\
                             &
                             \multicolumn{1}{|c|}{AM-GCN}               & - & $73.1$                & - \\ 
                             & \multicolumn{1}{|c|}{DGI}                  & $82.3_{\pm0.6}$ & $71.8_{\pm0.7}$ & $76.8_{\pm0.6}$ \\
                             &   \multicolumn{1}{|c|}{GRAND}               & $85.4_{\pm0.4}$ & $75.5_{\pm0.4}$                & $82.7_{\pm0.6}$ \\ 
                             & \multicolumn{1}{|c|}{Adaptive Weighting$\dagger$}               & $85.5_{\pm0.5}$ & $75.5_{\pm0.4}$                & $82.6_{\pm0.4}$ \\
                            & \multicolumn{1}{|c|}{Ensemble}  &              $85.6_{\pm0.4}$&                             $75.1_{\pm0.3}$ & $82.6_{\pm0.4}$           \\\hline
\multirow{1}{*}{Ours} 
                             & \multicolumn{1}{|c|}{AKE-GNN} &              $\bm{85.9_{\pm0.3}}$&                             $\bm{75.8_{\pm0.3}}$&
                             $\bm{83.0_{\pm0.6}}$
                             \\ \Xhline{0.2ex}
\end{tabular}
}
% \vspace{-0.5em}
\end{table}
\subsection{Extending AKE-GNN to Multiple GNNs} 
\label{sec:multiple-gnn}
AKE-GNN can be easily extended to the multiple GNN case, as illustrated in Fig.~\ref{fig:framework}(c). 
In each iteration of the knowledge exchange phase, each GNN model $\gG_k$ accepts the knowledge from $\gG_{k-1}$. After certain iterations of knowledge exchange, each GNN model contains the knowledge from all the other GNN models trained on the multiple graph views. 
We list the complete algorithm of AKE-GNN for multiple GNNs in Appendix.
\section{Experiments}
\label{sec:experiments}
\begin{table*}[ht]
\centering
\caption{Results of accuracy~(\%) on node classification tasks. $\Delta$ denotes absolute improvements between the backbone GNN model and AKE-GNN. The average values of $\Delta$ over all datasets are presented in brackets.}
\label{table:node}
\scalebox{0.82}{
\setlength{\tabcolsep}{11pt} % Default value: 6pt
\begin{tabular}{l ccc ccc ccc}
\toprule
                                                        & \textbf{Cora}                                                                            & \textbf{Citeseer}                                         & \textbf{Pubmed}                                           & \textbf{Chameleon}                                        & \textbf{Squirrel}                                         & \textbf{Actor}                                            & \textbf{Cornell}                                                  & \textbf{Texas}                                            & \textbf{Wisconsin}                                        \\ \midrule
\ \textbf{\# Nodes: $|\mathcal{V}|$}                                 & \multicolumn{1}{c}{2,708}                                                       & \multicolumn{1}{c}{3,327}                        & \multicolumn{1}{c}{19,717}                       & \multicolumn{1}{c}{2,277}                        & \multicolumn{1}{c}{5,201}                        & \multicolumn{1}{c}{7,600}                        & \multicolumn{1}{c}{183}                                  & \multicolumn{1}{c}{183}                          & \multicolumn{1}{c}{251}                          \\
\ \textbf{\# Edges: $|\mathcal{E}|$}                                 & \multicolumn{1}{c}{5,278}                                                       & \multicolumn{1}{c}{4,676}                        & \multicolumn{1}{c}{44,327}                       & \multicolumn{1}{c}{31,421}                       & \multicolumn{1}{c}{198,493}                      & \multicolumn{1}{c}{26,752}                       & \multicolumn{1}{c}{280}                                  & \multicolumn{1}{c}{295}                          & \multicolumn{1}{c}{466}                          \\
\ \textbf{\# Features: $\rd$}                                          & \multicolumn{1}{c}{1,433}                                                       & \multicolumn{1}{c}{3,703}                        & \multicolumn{1}{c}{500}                          & \multicolumn{1}{c}{2,325}                        & \multicolumn{1}{c}{2,089}                        & \multicolumn{1}{c}{931}                          & \multicolumn{1}{c}{1,703}                                & \multicolumn{1}{c}{1,703}                        & \multicolumn{1}{c}{1,703}                        \\
\ \textbf{\# Classes: $|\mathcal{Y}|$}                               & \multicolumn{1}{c}{7}                                                           & \multicolumn{1}{c}{6}                            & \multicolumn{1}{c}{3}                            & \multicolumn{1}{c}{5}                            & \multicolumn{1}{c}{5}                            & \multicolumn{1}{c}{5}                            & \multicolumn{1}{c}{5}                                    & \multicolumn{1}{c}{5}                            & \multicolumn{1}{c}{5}                            \\ \midrule\midrule
\textbf{GCN~\cite{kipf2016semi}}                                                     & $81.4_{\pm0.6}$                                                                 & $70.4_{\pm0.9}$                                  & $78.7_{\pm0.5}$                                  & $53.0_{\pm2.4}$                                  & $36.8_{\pm1.7}$                                  & $27.6_{\pm1.3}$                                  & $53.3_{\pm3.2}$                                          & $58.8_{\pm4.9}$                                  & $51.9_{\pm6.8}$                                  \\
\qquad+ FT                                              & $81.5_{\pm0.6}$                                                                 & $70.6_{\pm0.7}$                                  & $78.9_{\pm0.5}$                                  & $55.6_{\pm1.5}$                                  & $38.5_{\pm1.3}$                                  & $29.5_{\pm0.8}$                                  & $54.1_{\pm0.0}$                                          & $64.6_{\pm0.8}$                                  & $52.7_{\pm2.5}$                                  \\
% \qquad+ FT via single-view                                  & $81.9_{\pm0.5}$                                                                 & $69.9_{\pm0.6}$                                  & $79.2_{\pm0.3}$                                  & $56.2_{\pm1.2}$                                  & $40.1_{\pm0.1}$                                  & $30.3_{\pm0.6}$                                  & $54.1_{\pm0.0}$                                          & $71.4_{\pm3.2}$                                  & $54.1_{\pm1.3}$                                  \\
\qquad+ Ensemble                                   & $82.1_{\pm0.5}$                                                                 & $70.3_{\pm0.6}$                                  & $79.1_{\pm0.3}$                                  & $54.0_{\pm1.2}$                                  & $40.9_{\pm0.1}$                                  & $30.2_{\pm0.6}$                                  & $54.1_{\pm0.0}$                                          & $68.3_{\pm1.2}$                                  & $56.9_{\pm1.3}$                                  \\
\qquad+ Ensemble + FT                        & $82.2_{\pm0.4}$                                                                 & $70.5_{\pm0.3}$                                  & $79.4_{\pm0.5}$                                  & $56.2_{\pm0.8}$                                  & $40.8_{\pm0.2}$                                  & $30.6_{\pm0.5}$                                  & $54.1_{\pm0.0}$                                          & $68.5_{\pm1.8}$                                  & $56.9_{\pm1.2}$                                  \\
\qquad+ AKE-GNN                                      & $\bm{82.8_{\pm0.3}}$                                & $\bm{70.8_{\pm0.3}}$ & $\bm{79.6_{\pm0.1}}$ & $\bm{57.1_{\pm0.9}}$ & $\bm{41.4_{\pm0.1}}$ & $\bm{31.0_{\pm0.4}}$ & $\bm{54.1_{\pm0.0}}$         & $\bm{71.8_{\pm2.0}}$ & $\bm{57.5_{\pm0.9}}$ \\ \midrule
\multicolumn{1}{c}{\textbf{$\Delta$ \; (3.8)}} & \multicolumn{1}{c}{{\color[HTML]{FE0000} \textbf{1.4 $\uparrow$}}} & \multicolumn{1}{c}{\color[HTML]{FE0000}\textbf{0.3 $\uparrow$}}                 & \multicolumn{1}{c}{\color[HTML]{FE0000}\textbf{0.9 $\uparrow$}}                 & \multicolumn{1}{c}{\color[HTML]{FE0000}\textbf{4.1 $\uparrow$}}                 & \multicolumn{1}{c}{\color[HTML]{FE0000}\textbf{4.6 $\uparrow$}}                 & \multicolumn{1}{c}{\color[HTML]{FE0000}\textbf{3.4 $\uparrow$}}                 & \multicolumn{1}{c}{\color[HTML]{FE0000}\textbf{0.8 $\uparrow$}} & \multicolumn{1}{c}{\color[HTML]{FE0000}\textbf{13.0 $\uparrow$}}                & \multicolumn{1}{c}{\color[HTML]{FE0000}\textbf{5.6 $\uparrow$}}                 \\ \midrule\midrule
\textbf{GAT~\cite{velivckovic2017graph}}                                                     & $82.7_{\pm0.4}$                                                                 & $70.0_{\pm0.6}$                                  & $78.3_{\pm0.6}$                                  & $55.5_{\pm1.3}$                                  & $30.1_{\pm2.3}$                                  & $26.0_{\pm0.5}$                                  & $52.0_{\pm2.4}$                                          & $64.4_{\pm1.7}$                                  & $53.4_{\pm2.6}$                                  \\
\qquad+ FT                                              & $82.1_{\pm0.6}$                                                                 & $69.5_{\pm0.7}$                                  & $77.8_{\pm0.8}$                                  & $55.8_{\pm1.7}$                                  & $30.6_{\pm1.5}$                                  & $30.6_{\pm1.5}$                                  & $51.9_{\pm2.0}$                                          & $63.2_{\pm2.8}$                                  & $53.9_{\pm2.2}$                                  \\
% \qquad+ FT via single-view                                  & $81.8_{\pm0.8}$                                                                 & $69.7_{\pm0.6}$                                  & $78.4_{\pm0.6}$                                  & $56.5_{\pm1.5}$                                  & $32.2_{\pm1.1}$                                  & $32.2_{\pm1.1}$                                  & $54.1_{\pm0.0}$                                          & $63.2_{\pm1.3}$                                  & $56.3_{\pm2.0}$                                  \\
\qquad+ Ensemble                                   & $81.7_{\pm0.8}$                                                                 & $69.3_{\pm0.6}$                                  & $78.5_{\pm0.6}$                                  & $54.6_{\pm1.5}$                                  & $32.0_{\pm1.1}$                                  & $31.0_{\pm1.1}$                                  & $54.1_{\pm0.0}$                                          & $64.7_{\pm1.3}$                                  & $55.9_{\pm1.0}$                                  \\
\qquad+ Ensemble + FT                        & $82.7_{\pm0.3}$                                                                 & $69.5_{\pm0.4}$                                  & $78.5_{\pm0.2}$                                  & $56.8_{\pm0.5}$                                  & $32.0_{\pm0.8}$                                  & $32.1_{\pm0.8}$                                  & $54.0_{\pm1.1}$                                          & $64.8_{\pm0.0}$                                  & $55.8_{\pm1.8}$                                  \\
\qquad+ AKE-GNN                                      & $\bm{83.1_{\pm0.2}}$                                & $\bm{70.2_{\pm0.3}}$ & $\bm{78.7_{\pm0.4}}$ & $\bm{56.8_{\pm1.4}}$ & $\bm{32.5_{\pm0.2}}$ & $\bm{32.5_{\pm0.2}}$ & $\bm{54.1_{\pm0.1}}$         & $\bm{64.9_{\pm0.0}}$ & $\bm{56.9_{\pm1.6}}$ \\ \midrule
\multicolumn{1}{c}{\textbf{$\Delta$ \; (1.9)}} & \multicolumn{1}{c}{\color[HTML]{FE0000}\textbf{0.4 $\uparrow$}}                                                & \multicolumn{1}{c}{\color[HTML]{FE0000}\textbf{0.2 $\uparrow$}}                 & \multicolumn{1}{c}{\color[HTML]{FE0000}\textbf{0.4 $\uparrow$}}                 & \multicolumn{1}{c}{\color[HTML]{FE0000}\textbf{1.3 $\uparrow$}}                 & \multicolumn{1}{c}{\color[HTML]{FE0000}\textbf{2.4 $\uparrow$}}                 & \multicolumn{1}{c}{\color[HTML]{FE0000}\textbf{6.5 $\uparrow$}}                 & \multicolumn{1}{c}{\color[HTML]{FE0000}\textbf{2.1 $\uparrow$}}                         & \multicolumn{1}{c}{\color[HTML]{FE0000}\textbf{0.5 $\uparrow$}}                 & \multicolumn{1}{c}{\color[HTML]{FE0000}\textbf{3.5 $\uparrow$}}                 \\ \midrule\midrule
\textbf{APPNP~\cite{klicpera2018predict}}                                                   & $82.0_{\pm1.1}$                                                                 & $70.4_{\pm1.1}$                                  & $79.2_{\pm0.9}$                                  & $47.7_{\pm2.7}$                                  & $28.4_{\pm2.4}$                                  & $31.8_{\pm1.7}$                                  & $50.4_{\pm2.8}$                                          & $63.3_{\pm2.5}$                                  & $49.8_{\pm3.1}$                                  \\
\qquad+ FT                                              & $81.9_{\pm1.1}$                                                                 & $69.9_{\pm1.0}$                                  & $79.1_{\pm0.8}$                                  & $49.1_{\pm1.7}$                                  & $28.7_{\pm2.0}$                                  & $31.3_{\pm2.0}$                                  & $49.7_{\pm2.8}$                                          & $62.8_{\pm2.2}$                                  & $51.2_{\pm3.9}$                                  \\
% \qquad+ FT via single-view                                  & $82.0_{\pm0.6}$                                                                 & $69.5_{\pm0.8}$                                  & $78.5_{\pm1.1}$                                  & $49.5_{\pm2.1}$                                  & $30.0_{\pm0.7}$                                  & $31.2_{\pm1.2}$                                  & $50.3_{\pm2.5}$                                          & $63.0_{\pm2.4}$                                  & $51.6_{\pm2.8}$                                  \\
\qquad+ Ensemble                                   & $82.0_{\pm0.6}$                                                                 & $69.5_{\pm0.8}$                                  & $78.8_{\pm1.1}$                                  & $51.1_{\pm2.1}$                                  & $26.9_{\pm0.7}$                                  & $32.9_{\pm1.2}$                                  & $54.1_{\pm2.5}$                                          & $64.9_{\pm2.4}$                                  & $56.2_{\pm2.8}$                                  \\
\qquad+ Ensemble + FT                        & $82.3_{\pm0.7}$                                                                 & $69.5_{\pm1.3}$                                  & $79.7_{\pm0.5}$                                  & $51.7_{\pm1.5}$                                  & $31.4_{\pm0.8}$                                  & $32.9_{\pm0.8}$                                  & $51.0_{\pm1.8}$                                          & $64.7_{\pm1.9}$                                  & $56.0_{\pm1.2}$                                  \\
\qquad+ AKE-GNN                                      & $\bm{82.7_{\pm0.2}}$                                & $\bm{71.0_{\pm0.2}}$ & $\bm{79.7_{\pm0.1}}$ & $\bm{52.6_{\pm0.1}}$ & $\bm{31.6_{\pm1.3}}$ & $\bm{33.3_{\pm0.1}}$ & $\bm{55.4_{\pm1.4}}$         & $\bm{68.5_{\pm1.3}}$ & $\bm{56.9_{\pm0.1}}$ \\ \midrule
\multicolumn{1}{c}{\textbf{$\Delta$ \; (3.2)}} & \multicolumn{1}{c}{\color[HTML]{FE0000}\textbf{0.7 $\uparrow$}}                                                & \multicolumn{1}{c}{\color[HTML]{FE0000}\textbf{0.6 $\uparrow$}}                 & \multicolumn{1}{c}{\color[HTML]{FE0000}\textbf{0.5 $\uparrow$}}                 & \multicolumn{1}{c}{\color[HTML]{FE0000}\textbf{4.9 $\uparrow$}}                 & \multicolumn{1}{c}{\color[HTML]{FE0000}\textbf{3.2 $\uparrow$}}                 & \multicolumn{1}{c}{\color[HTML]{FE0000}\textbf{1.5 $\uparrow$}}                 & \multicolumn{1}{c}{\color[HTML]{FE0000}\textbf{5.0 $\uparrow$}}                         & \multicolumn{1}{c}{\color[HTML]{FE0000}\textbf{5.2 $\uparrow$}}                 & \multicolumn{1}{c}{\color[HTML]{FE0000}\textbf{7.1 $\uparrow$}}                 \\ \midrule\midrule
\textbf{JKNET-CAT~\cite{xu2018representation}}                                               & $81.9_{\pm0.5}$                                                                 & $69.2_{\pm0.5}$                                  & $78.3_{\pm0.2}$                                  & $55.7_{\pm2.2}$                                  & $36.4_{\pm1.7}$                                  & $28.5_{\pm1.3}$                                  & $53.9_{\pm4.7}$                                          & $59.5_{\pm5.1}$                                  & $51.5_{\pm5.4}$                                  \\
\qquad+ FT                                              & $81.3_{\pm0.4}$                                                                 & $68.1_{\pm1.0}$                                  & $77.8_{\pm0.6}$                                  & $56.9_{\pm1.4}$                                  & $36.1_{\pm0.9}$                                  & $28.3_{\pm0.5}$                                  & $52.7_{\pm2.5}$                                          & $61.9_{\pm0.8}$                                  & $51.4_{\pm2.1}$                                  \\
% \qquad+ FT via single-view                                  & $81.9_{\pm0.5}$                                                                 & $68.1_{\pm0.6}$                                  & $77.8_{\pm0.5}$                                  & $57.9_{\pm1.2}$                                  & $36.4_{\pm0.5}$                                  & $29.4_{\pm0.5}$                                  & $51.1_{\pm0.6}$                                          & $67.6_{\pm0.0}$                                  & $55.1_{\pm1.4}$                                  \\
\qquad+ Ensemble                                   & $82.0_{\pm0.5}$                                                                 & $67.2_{\pm0.6}$                                  & $77.6_{\pm0.5}$                                  & $58.1_{\pm1.2}$                                  & $36.6_{\pm0.5}$                                  & $29.5_{\pm0.5}$                                  & $54.1_{\pm0.6}$                                          & $67.6_{\pm0.0}$                                  & $56.9_{\pm1.4}$                                  \\
\qquad+ Ensemble + FT                        & $82.0_{\pm0.3}$                                                                 & $67.6_{\pm0.5}$                                  & $78.1_{\pm0.3}$                                  & $58.7_{\pm0.8}$                                  & $36.3_{\pm0.5}$                                  & $29.1_{\pm0.5}$                                  & $53.8_{\pm2.5}$                                          & $67.6_{\pm0.0}$                                  & $55.5_{\pm1.5}$                                  \\
\qquad+ AKE-GNN                                      & $\bm{82.4_{\pm0.2}}$                                & $\bm{69.5_{\pm0.1}}$ & $\bm{78.5_{\pm0.1}}$ & $\bm{59.4_{\pm0.1}}$ & $\bm{37.0_{\pm0.3}}$ & $\bm{29.6_{\pm0.4}}$ & $\bm{54.2_{\pm0.1}}$         & $\bm{67.6_{\pm0.0}}$ & $\bm{56.9_{\pm0.0}}$ \\ \midrule
\multicolumn{1}{c}{\textbf{$\Delta$ \; (3.2)}} & \multicolumn{1}{c}{\color[HTML]{FE0000}\textbf{0.5 $\uparrow$}}                                                & \multicolumn{1}{c}{\color[HTML]{FE0000}\textbf{0.3 $\uparrow$}}                 & \multicolumn{1}{c}{\color[HTML]{FE0000}\textbf{0.2 $\uparrow$}}                 & \multicolumn{1}{c}{\color[HTML]{FE0000}\textbf{3.7 $\uparrow$}}                 & \multicolumn{1}{c}{\color[HTML]{FE0000}\textbf{0.6 $\uparrow$}}                 & \multicolumn{1}{c}{\color[HTML]{FE0000}\textbf{1.0 $\uparrow$}}                 & \multicolumn{1}{c}{\color[HTML]{FE0000}\textbf{0.3 $\uparrow$}}                         & \multicolumn{1}{c}{\color[HTML]{FE0000}\textbf{8.1 $\uparrow$}}                 & \multicolumn{1}{c}{\color[HTML]{FE0000}\textbf{5.4 $\uparrow$}}                 \\ \midrule\midrule
\textbf{JKNET-MAX~\cite{xu2018representation}}                                               & $81.9_{\pm0.7}$                                                                 & $69.2_{\pm0.1}$                                  & $78.4_{\pm0.2}$                                  & $54.8_{\pm2.1}$                                  & $36.0_{\pm1.9}$                                  & $28.3_{\pm1.5}$                                  & $52.2_{\pm3.4}$                                          & $60.1_{\pm5.1}$                                  & $52.0_{\pm5.5}$                                  \\
\qquad+ FT                                              & $81.3_{\pm0.6}$                                                                 & $68.3_{\pm1.3}$                                  & $77.7_{\pm0.5}$                                  & $56.2_{\pm1.0}$                                  & $36.1_{\pm0.9}$                                  & $27.9_{\pm0.5}$                                  & $53.5_{\pm1.0}$                                          & $64.1_{\pm1.9}$                                  & $53.1_{\pm2.2}$                                  \\
% \qquad+ FT via single-view                                  & $81.7_{\pm0.6}$                                                                 & $68.4_{\pm0.6}$                                  & $78.2_{\pm0.3}$                                  & $56.5_{\pm1.2}$                                  & $36.5_{\pm0.8}$                                  & $28.2_{\pm0.8}$                                  & $54.1_{\pm0.0}$                                          & $65.4_{\pm1.4}$                                  & $55.1_{\pm2.0}$                                  \\
\qquad+ Ensemble                                   & $81.4_{\pm0.6}$                                                                 & $67.1_{\pm0.6}$                                  & $77.2_{\pm0.3}$                                  & $55.7_{\pm1.2}$                                  & $37.4_{\pm0.8}$                                  & $26.6_{\pm0.8}$                                  & $54.1_{\pm0.0}$                                          & $70.1_{\pm1.4}$                                  & $56.9_{\pm1.2}$                                  \\
\qquad+ Ensemble + FT                        & $82.3_{\pm0.2}$                                                                 & $67.7_{\pm0.5}$                                  & $77.8_{\pm0.3}$                                  & $56.6_{\pm1.0}$                                  & $36.4_{\pm0.5}$                                  & $28.0_{\pm0.5}$                                  & $54.1_{\pm0.0}$                                          & $66.9_{\pm1.2}$                                  & $55.3_{\pm0.8}$                                  \\
\qquad+ AKE-GNN                                      & $\bm{82.4_{\pm0.2}}$                                & $\bm{69.5_{\pm0.1}}$ & $\bm{78.6_{\pm0.1}}$ & $\bm{57.4_{\pm0.3}}$ & $\bm{37.7_{\pm0.2}}$ & $\bm{28.9_{\pm0.2}}$ & $\bm{54.1_{\pm0.0}}$         & $\bm{70.3_{\pm1.5}}$ & $\bm{57.6_{\pm1.0}}$ \\ \midrule
\multicolumn{1}{c}{\textbf{$\Delta$ \; (2.2)}} & \multicolumn{1}{c}{\color[HTML]{FE0000}\textbf{0.5 $\uparrow$}}                                                & \multicolumn{1}{c}{\color[HTML]{FE0000}\textbf{0.3 $\uparrow$}}                 & \multicolumn{1}{c}{\color[HTML]{FE0000}\textbf{0.2 $\uparrow$}}                 & \multicolumn{1}{c}{\color[HTML]{FE0000}\textbf{2.6 $\uparrow$}}                 & \multicolumn{1}{c}{\color[HTML]{FE0000}\textbf{1.7 $\uparrow$}}                 & \multicolumn{1}{c}{\color[HTML]{FE0000}\textbf{0.6 $\uparrow$}}                 & \multicolumn{1}{c}{\color[HTML]{FE0000}\textbf{1.9 $\uparrow$}}                         & \multicolumn{1}{c}{\color[HTML]{FE0000}\textbf{10.2 $\uparrow$}}                & \multicolumn{1}{c}{\color[HTML]{FE0000}\textbf{5.6 $\uparrow$}}                 \\ \midrule\midrule
\textbf{GCNII~\cite{chen2020simple}}                                                    & $85.4_{\pm0.4}$                                                                 & $73.2_{\pm0.5}$                                  & $80.0_{\pm0.3}$                                  & $60.7_{\pm2.1}$                                  & $44.1_{\pm1.9}$                                  & $30.1_{\pm0.9}$                                  & $61.2_{\pm2.1}$                                          & $67.0_{\pm6.3}$                                  & $74.7_{\pm3.4}$                                  \\
\qquad+ FT                                              & $84.8_{\pm0.5}$                                                                 & $72.1_{\pm0.8}$                                  & $78.1_{\pm2.3}$                                  & $61.9_{\pm1.6}$                                  & $45.0_{\pm0.9}$                                  & $30.0_{\pm0.6}$                                  & $60.9_{\pm2.5}$                                          & $68.6_{\pm1.8}$                                  & $72.7_{\pm3.4}$                                  \\
% \qquad+ FT via single-view                                  & $85.3_{\pm0.4}$                                                                 & $71.5_{\pm0.8}$                                  & $78.1_{\pm1.8}$                                  & $62.4_{\pm0.7}$                                  & $45.2_{\pm1.2}$                                  & $30.3_{\pm0.7}$                                  & $60.9_{\pm2.2}$                                          & $75.9_{\pm3.3}$                                  & $74.5_{\pm3.4}$                                  \\
\qquad+ Ensemble                                   & $85.2_{\pm0.3}$                                                                 & $72.8_{\pm0.6}$                                  & $78.4_{\pm1.8}$                                  & $62.9_{\pm0.7}$                                  & $43.8_{\pm1.2}$                                  & $30.3_{\pm0.7}$                                  & $59.5_{\pm2.2}$                                          & $78.4_{\pm3.3}$                                  & $72.6_{\pm3.4}$                                  \\
\qquad+ Ensemble + FT                        & $85.4_{\pm0.3}$                                                                 & $72.9_{\pm0.5}$                                  & $79.5_{\pm0.9}$                                  & $62.4_{\pm1.3}$                                  & $45.7_{\pm0.4}$                                  & $30.3_{\pm0.4}$                                  & $60.0_{\pm2.0}$                                          & $76.2_{\pm2.0}$                                  & $72.6_{\pm3.3}$                                  \\
\qquad+ AKE-GNN                                      & $\bm{85.7_{\pm0.3}}$                                & $\bm{73.8_{\pm0.3}}$ & $\bm{80.4_{\pm0.4}}$ & $\bm{63.1_{\pm1.3}}$ & $\bm{46.0_{\pm0.7}}$ & $\bm{31.4_{\pm0.1}}$ & $\bm{62.8_{\pm1.2}}$         & $\bm{79.7_{\pm2.8}}$ & $\bm{76.8_{\pm1.9}}$ \\ \midrule
\multicolumn{1}{c}{\textbf{$\Delta$ \; (2.6)}} & \multicolumn{1}{c}{\color[HTML]{FE0000}\textbf{0.3 $\uparrow$}}                                                & \multicolumn{1}{c}{\color[HTML]{FE0000}\textbf{0.6 $\uparrow$}}                 & \multicolumn{1}{c}{\color[HTML]{FE0000}\textbf{0.4 $\uparrow$}}                 & \multicolumn{1}{c}{\color[HTML]{FE0000}\textbf{2.4 $\uparrow$}}                 & \multicolumn{1}{c}{\color[HTML]{FE0000}\textbf{1.9 $\uparrow$}}                 & \multicolumn{1}{c}{\color[HTML]{FE0000}\textbf{1.3 $\uparrow$}}                 & \multicolumn{1}{c}{\color[HTML]{FE0000}\textbf{1.6 $\uparrow$}}                         & \multicolumn{1}{c}{\color[HTML]{FE0000}\textbf{12.7 $\uparrow$}}                & \multicolumn{1}{c}{\color[HTML]{FE0000}\textbf{2.1 $\uparrow$}}                 \\ 
% \midrule\midrule
\bottomrule
\end{tabular}
}
\end{table*}
\begin{table}[t]
\caption{Summary of results on four tasks. 
% $\Delta$ denotes absolute improvements between the backbone GNN model and AKE-GNN. The average values of $\Delta$ over all datasets are presented in brackets.
}
\vspace{-1.5em}
\begin{minipage}{1.0\linewidth}
\centering
\small
\vspace{1em}
\centerline{(a) Results of accuracy~(\%) on graph classification tasks. 
\vspace{0.5em}
% We use the default model selection strategies and data splits following~\cite{errica2019fair}.
}
\label{table:graph}
\setlength{\tabcolsep}{4.5pt}
\scalebox{0.84}{
\begin{tabular}{lccccccccc}
\toprule
                               & \textbf{DD}              & \textbf{NCI1}            & \textbf{PROTEINS}        & \textbf{IMDB}     & \textbf{REDDIT}    \\ \midrule
\textbf{GCN}                            & $76.5_{\pm3.0}$ & $76.3_{\pm1.4}$ & $73.3_{\pm3.9}$ & $70.5_{\pm4.5}$ & $88.6_{\pm2.3}$  \\
\qquad+ FT                       & $76.4_{\pm3.6}$ & $73.8_{\pm2.3}$ & $71.1_{\pm3.6}$ & $68.5_{\pm4.6}$ & $88.4_{\pm2.3}$  \\
\qquad+ Ensemble            & $79.5_{\pm2.5}$ & $80.5_{\pm1.2}$ & $75.9_{\pm3.2}$ & $64.8_{\pm4.3}$ & $81.5_{\pm3.1}$  \\
\qquad+ Ensemble + FT & $79.6_{\pm4.1}$ & $81.1_{\pm2.1}$ & $75.5_{\pm4.0}$ & $68.1_{\pm4.7}$ & $87.2_{\pm2.8}$  \\
\qquad+ AKE-GNN               & $\bm{81.0_{\pm3.4}}$ & $\bm{81.5_{\pm2.4}}$ & $\bm{80.4_{\pm3.5}}$ & $\bm{72.4_{\pm2.4}}$ & $\bm{89.5_{\pm1.9}}$  \\ \midrule
\qquad \textbf{$\Delta$ \; (3.9)}      & \color[HTML]{FE0000}\textbf{4.5 $\uparrow$}          & \color[HTML]{FE0000}\textbf{5.2 $\uparrow$}          & \color[HTML]{FE0000}\textbf{7.1 $\uparrow$}          & \color[HTML]{FE0000}\textbf{1.9 $\uparrow$}          & \color[HTML]{FE0000}\textbf{0.9 $\uparrow$}           \\ \midrule\midrule
\textbf{GIN}                            & $75.7_{\pm1.1}$ & $79.0_{\pm1.5}$ & $73.7_{\pm3.6}$ & $71.7_{\pm5.9}$ & $89.0_{\pm2.5}$  \\
\qquad+ FT                       & $69.2_{\pm3.3}$ & $74.1_{\pm4.0}$ & $66.3_{\pm4.3}$ & $68.5_{\pm7.5}$ & $81.1_{\pm10.3}$ \\
\qquad+ Ensemble            & $82.1_{\pm1.9}$ & $79.5_{\pm1.6}$ & $75.8_{\pm3.3}$ & $71.5_{\pm4.4}$ & $79.7_{\pm3.6}$  \\
\qquad+ Ensemble + FT & $81.0_{\pm3.5}$ & $79.8_{\pm2.0}$ & $70.7_{\pm4.3}$ & $70.5_{\pm5.6}$ & $88.7_{\pm4.4}$  \\
\qquad+ AKE-GNN               & $\bm{82.2_{\pm1.2}}$ & $\bm{79.8_{\pm2.0}}$ & $\bm{76.8_{\pm2.0}}$ & $\bm{72.3_{\pm6.3}}$ & $\bm{90.6_{\pm1.4}}$  \\ \midrule
\qquad \textbf{$\Delta$ \; (2.5)}      & \color[HTML]{FE0000}\textbf{6.5 $\uparrow$}          & \color[HTML]{FE0000}\textbf{0.8 $\uparrow$}          & \color[HTML]{FE0000}\textbf{3.1 $\uparrow$}          & \color[HTML]{FE0000}\textbf{0.6 $\uparrow$}          & \color[HTML]{FE0000}\textbf{1.6 $\uparrow$}           \\ 
% \midrule\midrule
\bottomrule
\end{tabular}
}
\end{minipage}
\vfill
\begin{minipage}{1.0\linewidth}
\vspace{1em}
\centerline{(b) Results of accuracy~(\%) on link prediction tasks.}
\vspace{0.5em}
\label{table:edge}
\scalebox{.79}{
\setlength{\tabcolsep}{3.5pt} % Default value: 6pt
\begin{tabular}{lcccccc}
\toprule
Dataset & \textbf{GCN} & FT & Ensemble & Ensemble + FT & AKE-GNN & \textbf{$\Delta$ (2.9)} \\
\midrule
\textbf{Cora}       & $90.3_{\pm1.2}$ & $91.6_{\pm0.9}$ 
& $93.8_{\pm0.6}$ & $94.3_{\pm0.7}$ & $\bm{94.6_{\pm0.8}}$
& \multicolumn{1}{c}{\color[HTML]{FE0000}\textbf{4.3 $\uparrow$}}\\
\textbf{CiteSeer}   & $89.1_{\pm1.3}$ & $89.8_{\pm0.8}$ 
& $92.1_{\pm0.7}$ & $92.1_{\pm0.6}$ & $\bm{92.6_{\pm0.4}}$
& \multicolumn{1}{c}{\color[HTML]{FE0000}\textbf{3.5 $\uparrow$}}\\
\textbf{PubMed}     & $95.4_{\pm0.3}$ & $95.6_{\pm0.2}$ 
& $95.3_{\pm0.2}$ & $96.1_{\pm0.2}$ & $\bm{96.2_{\pm0.2}}$
& \multicolumn{1}{c}{\color[HTML]{FE0000}\textbf{0.8 $\uparrow$}}\\
\bottomrule
\end{tabular}
}
\end{minipage}
\vfill
\begin{minipage}{1.0\linewidth}
\vspace{1em}
\centerline{(c) Results of accuracy~(\%) on OGBn-Arxiv.}
\vspace{0.5em}
\centering
\label{table:arxiv}
\setlength{\tabcolsep}{4pt}
\scalebox{0.8}{
\begin{tabular}{lccccccccc}
\toprule
\textbf{OGBn-Arxiv} 
& \textbf{GCNII} & \textbf{GCN\_res-v2} & \textbf{GCN\_DGL} & \textbf{GraphSAGE} \\
\midrule
\multirow{4}{*}{
\begin{tabular}{l}
\textbf{Original}\\
\quad + FT\\
% \quad + continue single-view\\
\quad + Ensemble\\
\quad + Ensemble + FT\\
\quad + AKE-GNN\\
\end{tabular}
}
& $72.7_{\pm0.2}$ & $71.5_{\pm0.9}$ & $70.9_{\pm0.5}$ & $71.5_{\pm0.4}$ \\
& $71.8_{\pm0.5}$ & $71.5_{\pm0.8}$ & $70.1_{\pm0.4}$ & $71.1_{\pm0.5}$  \\
& $71.8_{\pm0.3}$ & $71.9_{\pm0.4}$ & $70.4_{\pm0.3}$ & $71.4_{\pm0.4}$ \\
& $72.6_{\pm0.4}$ & $72.0_{\pm0.4}$ & $70.9_{\pm0.2}$ & $71.6_{\pm0.5}$  \\
& $\bm{73.3_{\pm0.6}}$ & $\bm{72.1_{\pm0.6}}$ & $\bm{71.5_{\pm0.7}}$ & $\bm{71.9_{\pm0.2}}$ \\
\midrule
\qquad \textbf{$\Delta$ \; (0.6)}
& {\color{red} \textbf{0.6 $\uparrow$} } 
& {\color{red} \textbf{0.6 $\uparrow$} } 
& {\color{red} \textbf{0.6 $\uparrow$} } 
& {\color{red} \textbf{0.4 $\uparrow$} }
\\
\bottomrule
\end{tabular}
}
\end{minipage}
\vspace{-1.8em}
\end{table}

\paragraph{Datasets \& tasks.} 
To show the generalization ability of the proposed AKE-GNN framework, we conduct experiments on 15 public benchmark datasets and 4 learning tasks.
(1) \textbf{Node classification}:
\emph{Citation network}~\cite{yang2016revisiting}: Cora, CiteSeer, and PubMed;
\emph{Wikipedia network}~\cite{rozemberczki2021multi}: Chameleon, and Squirrel; 
\emph{Actor co-occurrence network}~\cite{tang2009social}: Actor;
\emph{WebKB}~\cite{pei2020geom}: Cornell, Texas, and Wisconsin.
(2) \textbf{Link prediction}: \emph{Citation network}~\cite{yang2016revisiting}: Cora, CiteSeer, and PubMed.
(3) \textbf{Graph classification}: \emph{Chemical compounds}~\cite{Morris+2020}: DD, NCI1, and PROTEIN; \emph{Social network}~\cite{Morris+2020}: IMDB and REDDIT.
(4) \textbf{Large scale dataset}: Academic citation network in the open graph benchmark~(OGB)~\cite{hu2020open}: OGBn-Arxiv.
% Statistics of each dataset and experimental settings are summarized in Appendix G\&H.

\begin{figure*}[ht]
    \centering
    \includegraphics[width=1.\linewidth]{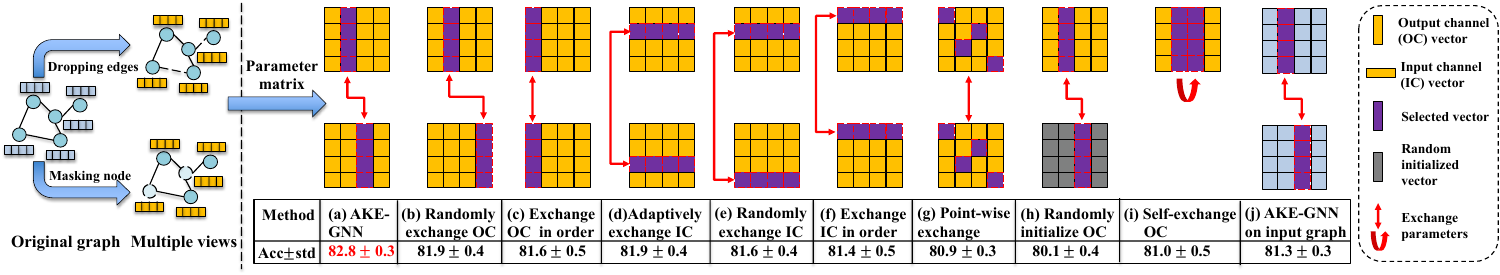}
    % \vspace{-1.5em}
    \caption{Comparison of different knowledge (weight matrix) exchange methods. We exchange the same number of parameters in the following methods for fair comparison and present the exchange scheme with one output/input channel for ease of understanding.~(a)~Adaptively exchange output channels (ours).~(b)~Randomly exchange output channels.~(c)~Exchange output channels in order.~(d)(e)(f)~Exchange input channels.~(g)~Exchange the weight matrix in a point-wise manner.~(h)~Randomly exchange output channels with a randomly initialized GNN.~(i)~Self-exchange output channels in the target GNN.~(j)~AKE-GNN without graph augmentations. 
    The experimental results are performed on Cora based on the GCN model. 
    }
    % \vspace{-1em}
    \label{fig:exchange}
\end{figure*}

\paragraph{Baselines.}
(1) \textbf{GNN backbone models}:
\whiteding{1} \emph{Node classification}: GCN~\cite{kipf2016semi}, GAT~\cite{velivckovic2017graph}, APPNP~\cite{klicpera2018predict}, JKNET with concatenation and maximum aggregation scheme~\cite{xu2018representation}, GRAND~\cite{feng2020graph}, and a recent deep GNN model GCNII~\cite{chen2020simple}.
\whiteding{2} \emph{Link prediction}: GCN.
\whiteding{3} \emph{Graph classification}: GCN, and GIN~\cite{xu2018powerful};
(2) \textbf{Weight re-activating methods}: Adaptive Weighting~\cite{meng2020filter}, and Decorr~\cite{jin2022feature};
(3) \textbf{Three variants for ablation analyses}:
\whiteding{1} Further training~(\textbf{FT}) trains GNN backbone models based on augmented graph views~(we report the best result among four graph views) with the same number of epochs as AKE-GNN to exclude the influence of longer training epochs and graph augmentations.
\whiteding{2} The multiple GNNs ensemble~(\textbf{Ensemble}) first trains GNNs on the generated views individually and then ensembles their outputs by majority voting.
\whiteding{3} The multiple GNNs ensemble+further training~(\textbf{Ensemble+FT}) not only ensembles the output of multiple GNNs on different views, but also trains each GNN with the same number of epochs as AKE-GNN.
The original GNN baseline models are denoted by their names directly.

\paragraph{Implementations.}
\label{sec:experiment-setup}
As a learning framework (rather than a specific GNN architecture), AKE-GNN is implemented \emph{based on} a backbone GNN model.
For generating multiple views, we adopt 4 graph augmentation methods, \ie~\emph{Masking node features}, \emph{Corrupting node features}, \emph{Dropping edges}, and \emph{Extracting subgraphs}.
% , which are detailed in Appendix B. 
We set $B=10$ as the number of bins in entropy calculation, $N=12$~(\#$\text{iterations}=3$) as the iteration steps, and $M=5$ as the number of exchange channels in each layer of GNNs.
Data preparation follows the standard experimental settings, including feature preprocessing and data splitting~\cite{feng2020graph,hu2020open}. 
We use accuracy (\%) with standard deviation averaged over 100 runs with different random seeds as the metric, except for the result on the large-scale \emph{OGBn-Arxiv} dataset, which is averaged over 10 runs.
Since each GNN in AKE-GNN interacts with all the other GNNs trained on different views, after the process of parameter exchange, the performance of different GNNs is similar. 
Thus, in what follows, we always record the performance of the first GNN model.
% Our codes and data are publicly available~\footnote{\url{https://tinyurl.com/AKE-GNN}}.

\subsection{Experimental Results}
\paragraph{Node classification.}
We implement AKE-GNN and the \textit{Ensemble} variant based on GRAND~\cite{feng2020graph}.
The comparison with baseline models on Cora, CiteSeer, and PubMed is reported in Table~\ref{table:results_of_public}.
% Detailed experimental results on node classification with more GNN backbone models can be found in Appendix C.
We find that AKE-GNN consistently outperforms both the single-view GNNs and multi-view GNNs, which demonstrates the effectiveness of adaptive knowledge exchange in modeling the relationship of multiple views.
Notably, AKE-GNN achieves the state-of-the-art results with 85.9\% accuracy on the Cora semi-supervised node classification dataset. AKE-GNN improves GCN by an average 3.9\% in terms of test accuracy on Cora.
% \footnote{\url{https://paperswithcode.com/sota/node-classification-on-cora-with-public-split}}.
The outperformance over the \emph{Ensemble} shows that the adaptive integration of multiple views in AKE-GNN is more effective than the simple ensemble of GNNs trained on different views.
Contrary to \emph{Ensemble}, which requires simultaneous inference of the multiple models, the inference cost of AKE-GNN is the same as a single-view model.
Moreover, we compare our methods with the adaptive weighting method following~\citet{meng2020filter}. 
% We find that 
AKE-GNN consistently surpasses this method, which suggests that our fine-grained method to exchange part of the knowledge in each GNN is more effective for multi-view GNNs.

To further evaluate the effectiveness of AKE-GNN, we implement AKE-GNN and compare it with the 3 variants of the baseline GNN model. The baseline models include GCN, GAT, APPNP, JKNet, and GCNII~\cite{kipf2016semi, velivckovic2017graph, klicpera2018predict, xu2018representation, chen2020simple}.
In Table~\ref{table:node}, the experimental results of baselines are reproduced based on their official codes.
% ~(see Appendix~\ref{sec:appendix-experiments} for more details).
% Experimental results are reported in Table~\ref{table:node}.
It shows that AKE-GNN consistently outperforms baselines by 1.9\%$\sim$3.8\% (absolute improvements) on average.
The outperformance of AKE-GNN over \emph{FT}, \emph{Ensemble}, and \emph{Ensemble+FT} shows that the expressiveness of the adaptive knowledge exchange comes from neither extra training epochs, nor the larger model capacity of multi-graph GNNs.
We find that AKE-GNN and \emph{Ensemble} outperform the baseline model by a large margin, which suggests the integration of multiple views on the small and medium-sized datasets helps to obtain better performance. 
% Note that in some cases, performing a certain augmentation~(dropping some important features) and longer training~(producing the over-fitting problem) may deteriorate the results. 
So, conducting \emph{FT} or \emph{Ensemble} among them brings inferior results. Nevertheless, AKE-GNN consistently achieves the best performance, which indicates its effectiveness to integrate informative information from multiple graph views.
Besides, in contrast to the ensemble methods, AKE-GNN utilizes only one network during inference, which is more computationally efficient.

\paragraph{Graph classification / link prediction.}
As shown in Table~\ref{table:graph}, AKE-GNN consistently outperforms the original GNN models by a large margin. 
Meanwhile, AKE-GNN achieves a higher accuracy over the three variants, further showing the superiority of our adaptive parameter exchange method. 
We notice that the performance improvement is marginal on the PubMed dataset of link prediction tasks. 
We postulate the reason is that the connection density~(\#edges / \#nodes)~of the PubMed dataset is higher than Cora and CiteSeer, which makes the model easier to complete the missing edges via message aggregation of neighbors in the single-view graph~\cite{pei2020geom}. 
Thus, AKE-GNN extracts less extra information behind the multiple generated graph views on PubMed than the other datasets, which hinders the model performance improvement.

\paragraph{Results on the large scale dataset.}
To validate that AKE-GNN can scale to large graphs, we further conduct experiments on the large citation dataset \emph{OGBn-Arxiv}.
We select four top-ranked GNN models from the leaderboard of OGB~\cite{hu2020open}, and then perform AKE-GNN based on them with the same GNN architectures and hyperparameters.
% ~(see Appendix H for more implementation details).
As shown in Table~\ref{table:arxiv}, our method outperforms the original methods and even their ensembles, which demonstrates the effectiveness of AKE-GNN on the large scale dataset. 

\begin{figure}[t]
    \centering
    \includegraphics[width=1.\linewidth]{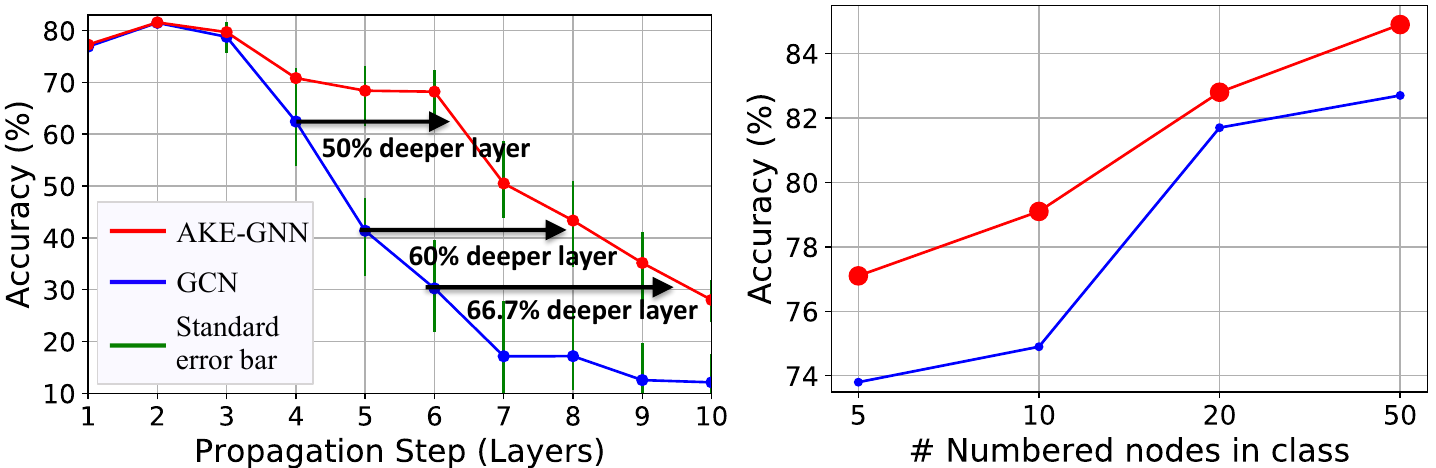}
    \caption{Performance comparisons of AKE-GNN with GCN on Cora. \textbf{Left:} the measurement of over-smoothing in terms of test accuracy~(\%). \textbf{Right:} test accuracy~(\%) on the few-shot setting.}
    % \vspace{-0.5em}
    \label{fig:over-smooth+few-shot}
\end{figure}
\subsection{Experimental Analyses}
\paragraph{Ablation study on knowledge exchange methods.}
\label{sec:exchange_method}
% We are particularly interested in 
%To verify the effectiveness of knowledge exchange methods in multi-view GNNs, 
%The dimension of parameter matrix is $\mathcal{R}^{C_l\times C_{l+1}}$, where the input channel is $C_l$ and the output channel is $C_{l+1}$. 
% Note that we exchange the same number of parameters in all the possible exchange methods for a fair comparison. 
% First, we make a comparison between two kinds of exchange approaches: 
% along the output channel (Figure~\ref{fig:exchange} (a)-(c)) and along the input channel (Figure~\ref{fig:exchange} (d)-(f)), where (a) (d), (b) (e) and (c) (f) utilize the adaptive exchange, random exchange, and orderly exchange respectively. 
% We first investigate whether maximum entropy is an effective metric to choose the output channel in the target network, and 
%A key question in AKE-GNN 
%is whether entropy maximization servers as the effective metric to choose the output channel in the target network. Thus, we first compare our adaptive knowledge exchange method (Fig.~\ref{fig:exchange} (a)) with two variants without the guidance of maximum entropy (Fig.~\ref{fig:exchange} (b)-(c)).
To verify the effectiveness of our adaptive exchange method, we compare AKE-GNN with other possible knowledge exchange approaches, as shown in Fig.~\ref{fig:exchange}.
These comparisons can be categorized into six groups: \whiteding{1} AKE-GNN~(a) v.s. output channel exchange~(b)(c), where (b) randomly chooses output channels in the target network and (c) swaps the first $M$ output channels; \whiteding{2} output channel exchange~(a)(b)(c) v.s. input channel exchange~(d)(e)(f); \whiteding{3} AKE-GNN (a) v.s. point-wise exchange (g); \whiteding{4} AKE-GNN~(a) v.s. adaptive exchange with a randomly initialized model~(h); \whiteding{5} AKE-GNN(a) v.s. adaptive exchange within the same network~(i); \whiteding{6} AKE-GNN~(a) v.s. AKE-GNN without graph augmentations~(j).

All the experiments are conducted on Cora using GCN~\cite{kipf2016semi} as the backbone model. 
For consistency, the same number of parameters are exchanged in all the experiments.
In Fig.~\ref{fig:exchange}, results are directly shown below the illustrations of the corresponding exchange methods.
We find that our proposed adaptive parameter exchange method along the output channel consistently outperforms all other approaches. 
From Fig.~\ref{fig:exchange}, we can conclude that: 
1) Comparing (a) with (b) and (c), the adaptive approach is more effective to substitute the redundant channel as it uses the entropy maximization heuristic;
2) Comparing (a) (b) (c) and (d) (e) (f), we find that exchanging the output channel is more effective than exchanging the input one.
As illustrated in Fig.~\ref{fig:tensor-multi}, each output hidden feature is solely determined by the corresponding output channel's parameters in the last layer. Thus, such a scheme may exchange the extra information from the other weight parameter matrix of GNN models, which enriches the model's representation ability;  
3) The result in (g) suggests that random point-wise exchange without considering the input or output channel decreases accuracy; 
4) Comparing (a) with (h), we can properly draw the conclusion that exchanging parameters with another well-trained GNN can incorporate knowledge of another graph view, and thus achieve better performance; 
5) Self-exchanging cannot bring benefits as shown in (i) since it does introduce extra knowledge;
6) Since exchanging knowledge without graph augmentations cannot introduce external information from other graph views, it achieves similar results as the baseline GCN and performs much worse than AKE-GNN, as shown in (j).
\begin{figure}[t]
    \centering
    \includegraphics[width=1.\linewidth]{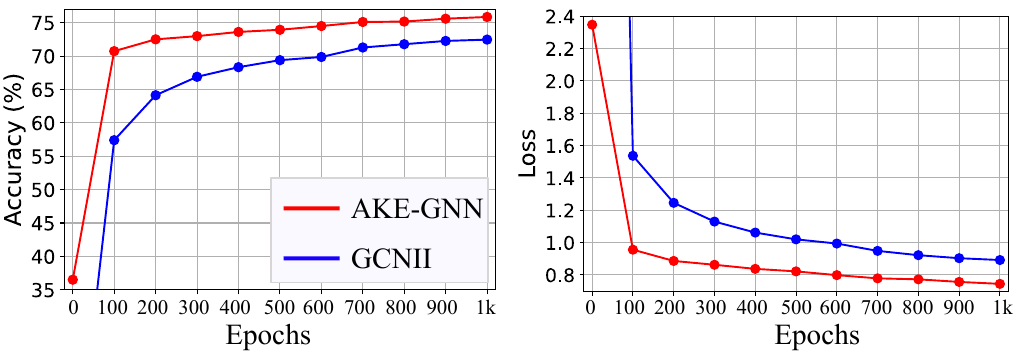}
    \caption{Learning curves (accuracy~(\textbf{left}) and loss~(\textbf{right})) on OGBn-Arxiv with the GCNII backbone model.}
    \label{fig:acc-loss1}
    % \vspace{-0.5em}
\end{figure}

\begin{figure*}[t]
    \centering
    \includegraphics[width=0.98\linewidth]{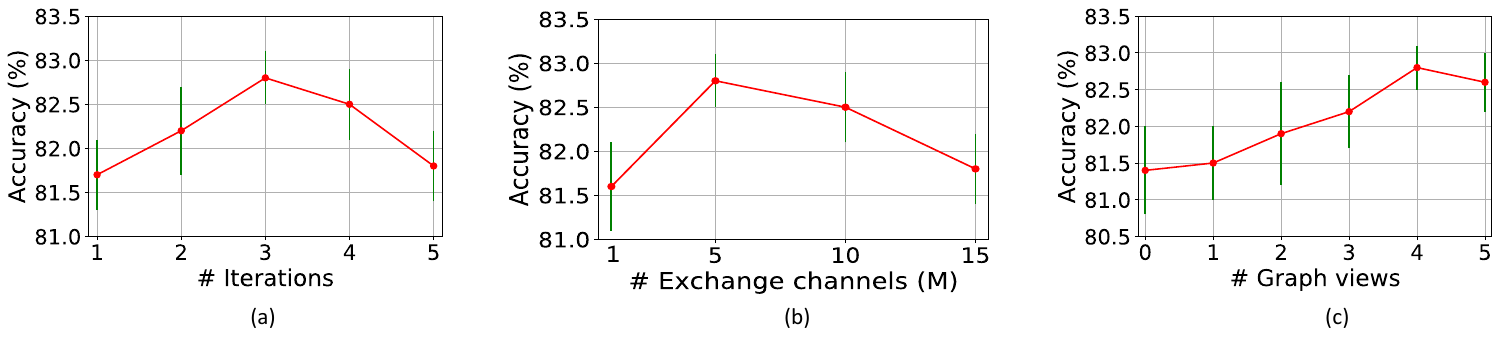}
    \caption{Hyperparameter analyses on the Cora dataset.}
    \label{fig:parameter-study}
\end{figure*}
\paragraph{Over-smoothing.}
Most current GNN models are shallow due to the over-smoothing issue, where node features become indistinguishable as we increase the feature propagation steps~\cite{liu2020towards}.
We present the results of GCN~\cite{kipf2016semi} by increasing the propagation steps (layers), and implement AKE-GNN based on GCN for comparison. In Fig.~\ref{fig:over-smooth+few-shot}, we empirically find that AKE-GNN can mitigate the over-smoothing issue compared to the original GCN. 
As the number of layers increases, the accuracy of the original GNN decreases dramatically from 0.8 to 0.1. 
In contrast, the accuracy of AKE-GNN decreases much slower. We find that AKE-GNN can make the propagation layer at least 50\% deeper~(propagation layer from 4 to 10) than the original GCN model without sacrificing the learning performance.
This suggests that AKE-GNN equipped with the adaptive knowledge exchange method provides an effective way to extend model capacity with relatively large layer numbers.
As suggested in~\cite{rong2019dropedge}, dropping edges in some generated graph views may help mitigate over-smoothing issues. We conjecture that removing certain edges makes node connections more sparse, and hence avoids over-smoothing to some extent when AKE-GNN goes very deep.

\paragraph{Few-shot.}
Following prior work~\cite{wan2021contrastive}, we further evaluate the effectiveness of AKE-GNN under the few-shot setting.
Taking Cora as the representative dataset, we manually vary the number of labeled nodes per class from 1 to 50 in the training phase, and keep the validation and test dataset unchanged. As shown in Fig.~\ref{fig:over-smooth+few-shot}, AKE-GNN consistently outperforms GCN.
Specifically, the relative improvements on accuracy are 4.0/3.3/4.2/0.9/2.2 on average for 1/5/10/20/50 labeled nodes per class, which shows that exchanging information from multiple generated graph views is more efficient when utilizing limited supervision.

\paragraph{Accuracy and loss curves.}
We plot the training loss and accuracy curves to verify that AKE-GNN can improve the training process compared with the backbone GNN model~(GCNII).
Fig~\ref{fig:acc-loss1} shows the accuracy and the loss curve of AKE-GNN in the re-training phase and GCNII on \emph{OGBn-Arxiv}.
We can observe that the blue line~(AKE-GNN) is above the orange line~(GCNII) in Fig.~\ref{fig:acc-loss1}(a), while the blue line~(AKE-GNN) is under the orange line~(GCNII) in Fig.~\ref{fig:acc-loss1}(b).
It demonstrates that the backbone GNN model equipped with our proposed AKE-GNN framework indeed converges faster.

\paragraph{Hyperparameter study.}
% We study the sensitivity of hyperparameters in AKE-GNN, and more details could be found in Appendix D.
% In all, the performance of our framework is relatively stable across different hyperparameters and thus does not rely on heavy and case-by-case hyperparameter tuning.
We study the sensitivity of hyperparameters of our framework AKE-GNN, and conduct experiments on Cora based on the GCN model. We have two hyperparameters in the knowledge exchange phase of AKE-GNN: the iteration steps $N$ and the exchanging channels $M$. 
Taking AKE-GNN on 4 graph views as an example, we first present a study on the number of iterations by varying it from $1$ to $5$~($N = 4 \times \#$Iterations) while using the default value $M=5$. As shown in Fig.~\ref{fig:parameter-study} (a), adaptively exchanging parameters with only a few iterations~($\#$Iterations=3) can achieve satisfying performance.
We further study the number of exchange channels $M$ by varying it from 1 to 15 (the hidden size of the GCN is 16) while fixing $\#$Iterations=3 in Fig.~\ref{fig:parameter-study} (b). 
The best performance is achieved by exchanging part of channels rather than all parameters in a layer, demonstrating that adaptively exchanging information in a complementary way can bring more benefits.
Finally, we study the number of graph views from 1 to 5 while using the default value $\#$Iterations=3 and $M=5$. We successively add the following graph views in AKE-GNN: \emph{masking node features}, \emph{corrupting node features}, \emph{dropping edges}, \emph{extracting subgraphs}, \emph{the original graph}. As shown in Fig.~\ref{fig:parameter-study} (c), adding more graph views indeed improves performance. However, the performance stagnates when we add the number of graph views to 5. We assume the cause might be that the network receives too much information from other graph views which may affect its self-information for learning.
In all, we find that the performance of our framework is relatively stable across different hyperparameters, and thus does not rely on heavy and case-by-case hyperparameter tuning to achieve satisfactory results. 

\paragraph{Discussion of computational complexity.}
The computational complexity of AKE-GNN is $O\left(NLd^3\right)$, where $N$ is the number of iterations, $L$ is the number of GNN layers, and $d$ is the embedding dimension. Note that $N$ is usually small, and we set 3 in our experiments. The time cost of AKE-GNN is acceptable compared with multi-epoch training of GNNs, because the adaptive parameter exchange only executes once before the re-training of GNNs. 
% The time complexity of a running example could be found in Appendix E.
We conduct ablations on the large paper citation network~(\emph{OGBn-Arxiv}) using GCNII~\cite{chen2020simple} to investigate the time consumption overhead of AKE-GNN. We set the hidden dimension sizes as 64, 128, and 256. As shown in Table~\ref{tab:time}, the time cost of the adaptive parameter exchange is substantially less than that in the training phase, which indicates that the bottleneck of AKE-GNN still depends on the training of GNNs rather than the adaptive parameter exchange.
\begin{table}[t]
\caption{Time consumption overhead in terms of seconds with different hidden dimension sizes.}
\label{tab:time}
\centering
\scalebox{1.}{
\setlength{\tabcolsep}{7.5pt}
\begin{tabular}{lcccc}
\toprule
Training phase/Hidden size               & \textbf{64} & \textbf{128} & \textbf{256} & \\ \midrule
GCNII~\cite{chen2020simple}        & 1,170                             & 2,220                               & 4,200                               & \\
\textbf{Adaptive exchange} & 228                               & 575                                & 1,524                               & \\
Re-training          & 1,149                              & 2,139                               & 4,546                               & \\
    \bottomrule
\end{tabular}
}
\vspace{-1em}
\end{table}
\section{Conclusion and Future Work}
\label{sec:conclusion}
In this paper, we propose a novel framework named AKE-GNN, which performs the adaptive knowledge exchange strategy on the multiple GNNs each corresponding to a generated graph view. In AKE-GNN, we iteratively exchange redundant channels in the weight matrix of one GNN with informative channels of another GNN in a layer-wise manner. 
% % AKE-GNN does not rely on GNN architectures and approaches to generating multiple views. Thus, 
Moreover, existing GNNs can be seamlessly integrated into our framework. Comprehensive experiments show that our proposed learning framework consistently outperforms the existing popular GNN models and even their ensembles. This work focuses on exchanging knowledge of views with different graph augmentations. However, how to generate diverse views for better knowledge exchange is still under exploration, which leaves for future work.
We hope our work can inspire new ideas in exploring new learning mechanisms on the multi-view graphs.

\section*{Acknowledgments}
Liang Zeng, Jin Xu, and Jian Li are supported in part by the National Natural Science Foundation of China Grant 62161146004.

\clearpage
\balance
\bibliographystyle{ACM-Reference-Format}
\bibliography{cikm23}
% \clearpage
% \appendix
% \input{AppendixA-algorithm}
% \input{AppendixD-datasets}
% \input{AppendixE-experiments}

\end{document}